\PassOptionsToPackage{table,xcdraw}{xcolor}
\documentclass[sigconf]{acmart}

\AtBeginDocument{%
  }

\setcopyright{acmlicensed}
\copyrightyear{2018}
\acmYear{2018}
\acmDOI{XXXXXXX.XXXXXXX}
\acmISBN{978-1-4503-XXXX-X/2018/06}

\usepackage{amssymb}
\usepackage{tcolorbox} 
\usepackage{graphicx}
\usepackage{float}
\usepackage{multirow}
\usepackage{array}
\usepackage{arydshln}
\usepackage[utf8]{inputenc}
\definecolor{myyellow}{RGB}{230,204,180}
\usepackage{amsmath}
\usepackage{centernot}
\usepackage{makecell}

\usepackage{tikz}
\usetikzlibrary{arrows.meta}

\usepackage{enumitem}
\begin{document}

\title{Prompting Large Language Models with Partial Knowledge for Answering Questions with Unseen Entities}

\author{Zhichao Yan}
\affiliation{%
  \institution{School of Computer and Information Technology, Shanxi University}
  \city{Taiyuan}
  \country{China}}
\email{202312407023@email.sxu.edu.cn}

\author{Jiapu Wang}
\affiliation{%
  \institution{Beijing University of Technology}
  \city{Beijing}
  \country{China}}
\email{jiapuwang9@gmail.com}

\author{Jiaoyan Chen}
\affiliation{%
  \institution{Department of Computer Science, University of Manchester}
  \city{Manchester}
  \country{England}}
\email{jiaoyan.chen@manchester.ac.uk}

\author{Yanyan Wang}
\affiliation{%
  \institution{School of Computer and Information Technology, Shanxi University}
  \city{Taiyuan}
  \country{China}}
\email{yanyanwang@sxu.edu.cn}

\author{Hongye Tan}
\affiliation{%
  \institution{School of Computer and Information Technology, Shanxi University}
  \city{Taiyuan}
  \country{China}}
\email{tanhongye@sxu.edu.cn}

\author{Jiye Liang}
\affiliation{%
  \institution{School of Computer and Information Technology, Shanxi University}
  \city{Taiyuan}
  \country{China}}
\email{ljy@sxu.edu.cn}

\author{Xiaoli Li}
\affiliation{%
  \institution{Singapore University of Technology and Design}
  \country{Singapore}
}
\email{xiaoli_li@sutd.edu.sg}

\author{Ru Li}
\affiliation{%
  \institution{School of Computer and Information Technology, Shanxi University}
  \city{Taiyuan}
  \country{China}}
\email{liru@sxu.edu.cn}

\author{Jeff Z.Pan}

\affiliation{%
  \institution{ILCC, School of Informatics, University of Edinburgh}
  \city{Edinburgh}
  \country{UK}}
\email{j.z.pan@ed.ac.uk}

\renewcommand{\shortauthors}{Yan et al.}


\begin{abstract}
Retrieval-Augmented Generation (RAG) shows impressive performance by supplementing and substituting parametric knowledge in Large Language Models (LLMs). Retrieved knowledge can be divided into three types: explicit answer evidence, implicit answer clue, and insufficient answer context  which can be further categorized into totally irrelevant and partially relevant  information.  Effectively utilizing partially relevant knowledge remains a key challenge for RAG systems, especially in incomplete knowledge base retrieval. Contrary to the conventional view, we propose a new perspective: LLMs can be awakened via partially relevant knowledge  already embedded in LLMs. To comprehensively investigate this phenomenon, the triplets located in the gold reasoning path and their variants are used to construct partially relevant knowledge by removing the path that contains the answer. We provide theoretical analysis of the awakening effect in LLMs and support our hypothesis with experiments on two Knowledge Graphs (KGs) Question Answering (QA) datasets. Furthermore, we present a new task, Unseen Entity KGQA, simulating real-world challenges where entity linking fails due to KG incompleteness. Our awakening-based approach demonstrates greater efficacy in practical applications, outperforms traditional methods that rely on embedding-based similarity which are prone to returning noisy information.
\end{abstract}




\maketitle

\section{Introduction}

\begin{figure}[t]
    \centering
    \includegraphics[scale=0.38]{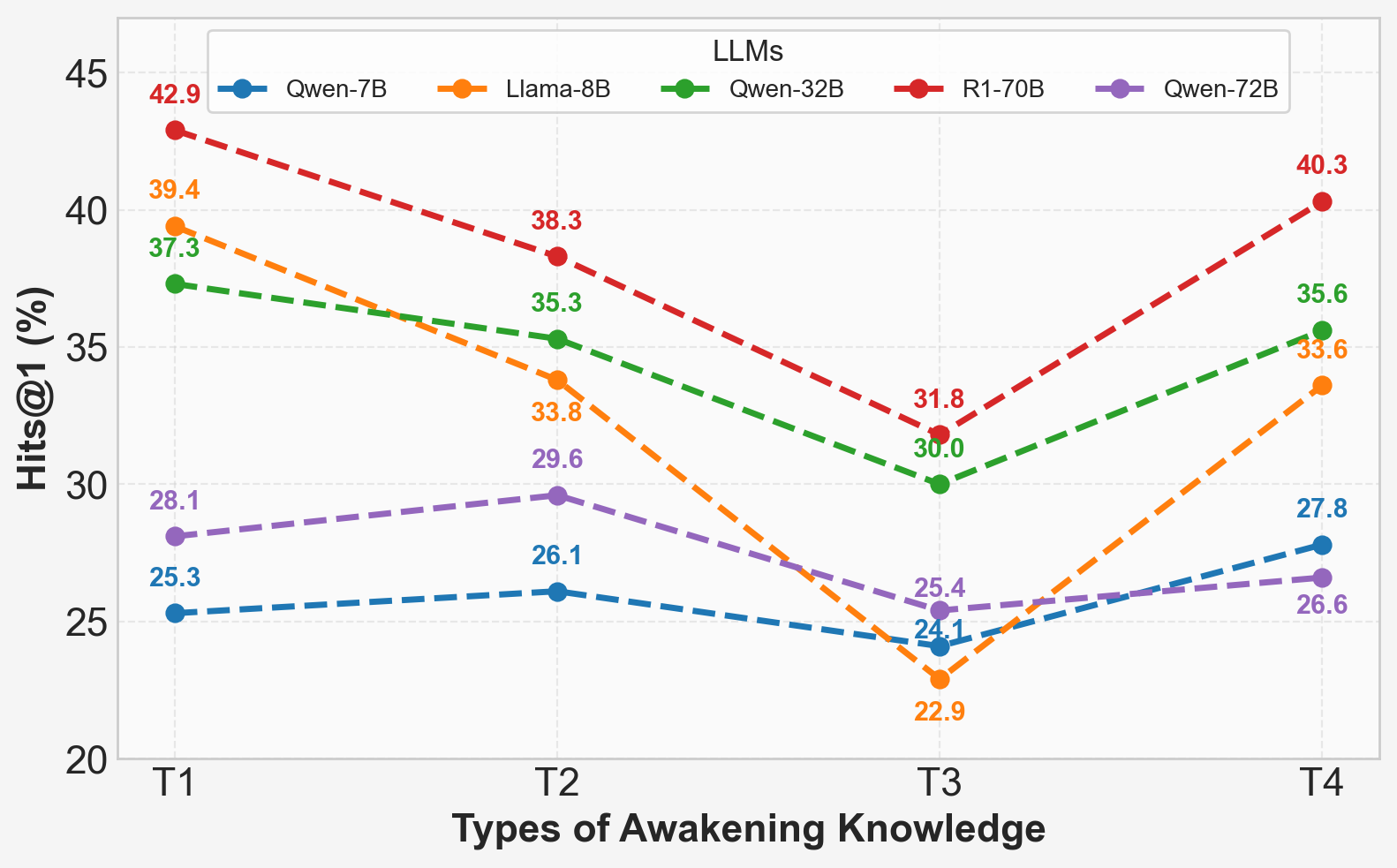}
    \caption{
    The KGQA performance (Hits@1) of LLMs when fed with different types of awakening knowledge, tested on questions that cannot be answered by the LLMs alone. T1 is the start entity (mentioned in question) with gold outgoing relation in reasoning path; T2 is an arbitrary triple in the KG that has the same start entity as T1 and non-gold outgoing relation; T3 is the inverse of T2, consisting of incoming relation and arbitrary entity. T4 is the type of the start entity. More kinds of awakening knowledge are introduced in Section \ref{Benchmarking}.}
    
    \Description{}
    \label{fig:heatmap}
\end{figure}

Large Language Models (LLMs) \cite{10387715, brown2020language, wang2024large} have achieved remarkable success across a wide range of natural language tasks, driven by pre-training on massive text corpora. Despite their impressive capabilities, LLMs often suffer from outdated knowledge and hallucinations, which undermine their reliability in real-world applications. These limitations stem from the static nature of their parametric memory which cannot dynamically incorporate 
newly emerging facts or verify information during inference.

Retrieval-Augmented Generation (RAG) \cite{lewis2020retrieval, gao2023retrievalaugmented, fan2024survey, gao2023enabling} has emerged as a powerful framework that integrates external knowledge into the generation process. By retrieving relevant information at inference time, RAG expands the knowledge capacity of LLMs beyond their fixed parameters, mitigating hallucinations and improving factual consistency \cite{asai2023self, soudani2024fine, joren2025sufficient}. The retrieved knowledge in RAG can be broadly categorized into three types \cite{yoranmaking, cuconasu2024power, joren2025sufficient,amiraz-etal-2025-distracting}:
(I) Explicit Answer Evidence, where the retrieved content directly contains the correct answer;
(II) Implicit Answer Clue, where the answer is not explicitly stated but can be inferred through reasoning; and
(III) Insufficient Answer Context, where the retrieved content cannot independently support an answer. The third category can be further divided into
\textit{(a) Totally Irrelevant}, which has no semantic connection to the question, and
\textit{(b) Partially Relevant}, which contains related contexts but is insufficient for complete inference.

 \begin{figure*}[t]
    \centering
\includegraphics[scale=0.093]{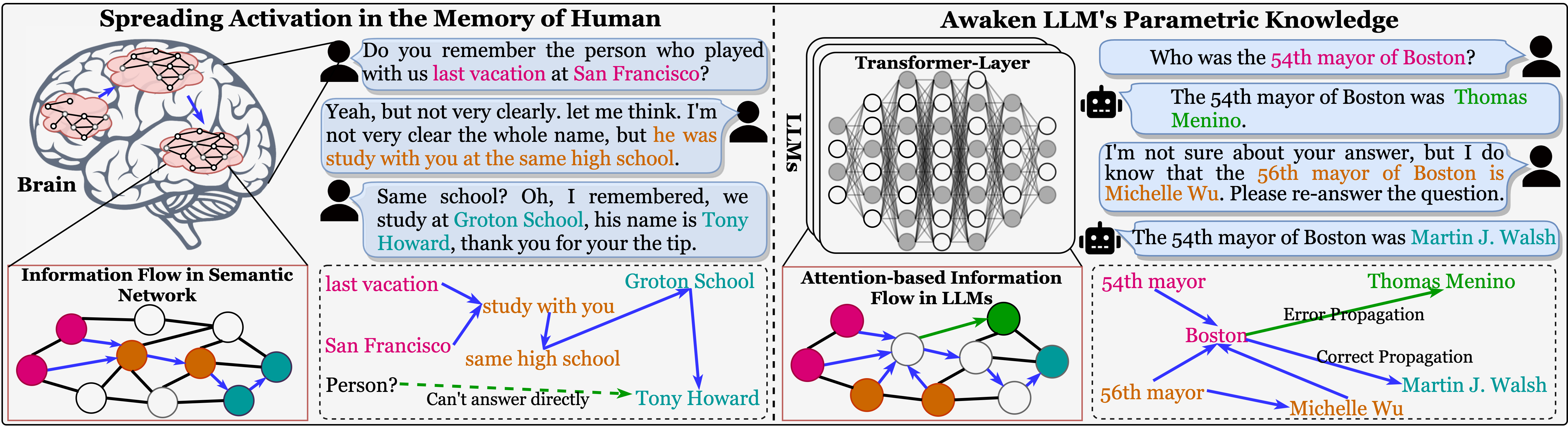}
    \caption{Motivation for Awakening Large Language Models. Human can recall the relevant knowledge ``Tony Howard'' via the existence knowledge ``same high school'' and ``Groton School'' spreading in Semantic Network \cite{anderson1983spreading}. Similarly, LLMs are awakened by partially relevant knowledge ``56th mayor of Boston is Michelle Wu'', which shares a similar relation with ``54th mayor of Boston'' \cite{dai2022knowledge, jimenez2024hipporag}.  This connection is established through the spreading activation, facilitating the propagation of information to the ``Martin J. Walsh''.}
    \Description{Illustration depicting 
    the analogy between human brain activation and large language model knowledge awakening.}
    \label{fig:introduction}
\end{figure*}

Prior research has primarily focused on leveraging explicitly retrieved knowledge, yet effectively activating and utilizing partially relevant knowledge remains a significant challenge. Traditional RAG methods typically inject fully relevant external knowledge to either augment 
the capacity of LLMs or replace outdated parametric information. However, as demonstrated in Figure \ref{fig:heatmap}, when partially relevant knowledge is already embedded within an LLM and reintroduced as contextual input, it reactivates latent internal representations that otherwise remain inaccessible. 

This phenomenon closely mirrors the Spreading Activation Theory of Memory \cite{anderson1983spreading, anderson2013architecture}, which posits that human memory is organized as a semantic network of \textit{Cognitive Units}. When a specific unit is activated by a cue, activation propagates to semantically related units, thereby strengthening interconnections and guiding the flow of retrieval. This mechanism enhances the recall of vague or partially encoded knowledge and mitigates the failure to access already learned information (Figure~1, left). Drawing parallels to LLMs, Dai \textit{et al.} \cite{dai2022knowledge} observed that activation patterns within transformer models are positively correlated with knowledge sharing similar relational structures. Such internal activations facilitate latent relational generalization and echo the associative retrieval dynamics observed 
in human memory (Figure~1, right).

To comprehensively investigate the phenomenon, we utilize a Knowledge Graph (KG) to verify and analyze what type of knowledge can awaken an LLM, which we refer to as awakening knowledge. Briefly, we initially choose questions that cannot be answered by LLMs alone without additional input, and then for each question, we employ partial facts \(\mathcal{K}\) from the complete knowledge which are the path that can infer the answer in the KG. Specifically, the knowledge in the gold reasoning path comprises entities and relations, including the start, intermediate, and end (answer) entities. Each entity is associated not only with gold incoming and outgoing relations on the path but also with non-gold relations and type information. Due to the absence of either incoming or outgoing gold relations, the start and end entities involve only four relation types. Beyond \(\mathcal{K}\), its variant knowledge (also denoted as \(\mathcal{K}\)) consists of any combination of entities and relations from the KG, where either the entity or the relation must originate from the gold path. (Refer Table \ref{tab:type}.)
We then conduct knowledge probing to select facts from \(\mathcal{K}\)  that are contained in LLMs, and further probe whether these selected facts can awaken LLMs, i.e., whether they can enable LLMs to correctly answer the question when fed into LLMs. Experimental results on two KGQA datasets --- 2WikiMultiHop (2Wiki) and ComplexWebQuestions (CWQ) demonstrate that all examined LLMs exhibit knowledge awakening to varying degrees, which means LLMs do not always make optimal use of the knowledge encoded in their parameters, and the explicit reintegration and activation of relevant knowledge can significantly enhance their performance.

We also find the phenomena of awakening LLM internal knowledge can be utilised for augmenting RAG. 
We propose a realistic and common RAG scenario called \textbf{Unseen Entity KGQA}, where the entity mentioned in a question does not have corresponding entity in the KG, construct two datasets of this scenario using the whole KGs of Wikidata and Freebase as well as the questions of Mintaka, and finally develop and extensively evaluate a new method that uses LLM awakening rules to selectively extract different kinds of partial knowledge in the KG to awaken the LLM.

Overall, our contribution can be summarized as follows:
\begin{itemize}[left=0cm]
    \item We systematically study the phenomenon of LLM knowledge awakening, which indicates that prompting an LLM with a part of the knowledge it has encoded can still significantly enhance its performance via structured latent activation, with a theoretical analysis of the occurrence mechanisms and extensive experiments for validation.
    \item We propose a new RAG task named Unseen Entity KGQA, which assumes the KG is incomplete with no entities matched to the questions' head entities, and accordingly develop two datasets with the original complete KGs of Wikidata and Freebase, so as to demonstrate the potential usage of the LLM awakening phenomenon.
    \item We develop a new KGQA method based on selective knowledge retrieval with the LLM awakening rules, and it achieves promising results on the three Unseen Entity KGQA datasets. 
\end{itemize}

\section{Related Work}
We review two main research areas to improve the performance of LLMs. First, much work utilizes the query more effectively to retrieve relevant information. Another line of study is prompting LLMs to better utilize parametric knowledge.

\subsection{Retrieval Augmented Generation}
According to the representation form of text knowledge, Mainstream RAG research has centered on document retrieval, KG retrieval and hybrid strategies that combine the two.

\textbf{Document retrieval.} With the deep study in RAG, the primary bottleneck in this area is retrieval noise. This noise appears unavoidable given current technological capabilities \cite{cuconasu2024power, fang-etal-2024-enhancing, wu2024how, liu2025robust}. Query2doc \cite{wang2023query2doc} employs LLMs to expand queries.  Bhunia \textit{et al.} \cite{bhunia2022sketching} propose a reinforcement learning-based stroke selection method that enhances noise robustness and retrieval performance by adaptively retaining strokes according to their contribution. RAAT \cite{fang-etal-2024-enhancing} trains models using adaptive adversarial training to improve robustness against retrieval noise. Ret-Robust \cite{yoranmaking} first employs a natural language inference model to filter irrelevant passages, and then trains on a mixture of relevant and irrelevant contexts to enhance robustness. Joren \textit{et al.} \cite{joren2025sufficient} present a selective generation method that uses sufficient context information to guide abstention. Document-level retrieval inevitably brings noise information and lacks structured reasoning path. 

\textbf{KG retrieval.} To address the above challenges, Wang \textit{et al.} \cite{wang-etal-2021-retrieval} retrieve knowledge from subgraphs that are connected via entity link models. G-retriever \cite{he2024g} employs graph neural networks as a retriever to extract knowledge from subgraphs. ToG \cite{sunthink} links the entities in the question with KGs and generates reasoning pathways to identify promising paths. RoG \cite{luoreasoning} uses LLMs for planning and performs reasoning on retrieved triplets from subgraphs, to enhance the faithfulness. However, subgraphs cannot always be extracted because of the incompleteness which limits the applicability of these methods in real-world scenarios. 

To bridge the gap between document retrieval and KG retrieval, GraphRAG \cite{edge2024local} constructs a KG from the documents and then retrieves information from the graph after community detection. LightRAG \cite{guo2024lightrag} cuts the community detection process in GraphRAG and utilizes a dual-level retrieval paradigm to enhance the retrieval.

\textbf{Reasoning with Retrieval.} 
Some recent papers have noticed that the intrinsic limitations of LLMs cannot be further alleviated by long-chain reasoning \cite{li2025search, shah2025rethinking, gandhi2025cognitive}. Test-time scaling with retrieval is becoming popular due to the success of DeepSeek-R1 and GPT-o1 \cite{li2025search, song2025r1, jin2025search, chen2025learning}. The methods utilize reinforcement learning to train LLMs and generate retrieval tokens when retrieval is needed. Liao \textit{et al.} \cite{liao2025awakening} propose AAG that tries to awaken the internal knowledge of LLMs via retrieved top-k similar documents, which is close to the traditional RAG. Different from AAG, we utilize the knowledge contained in LLMs to awaken them, identifying when the knowledge is insufficient to answer the question.

\subsection{Prompting LLMs}

Much work focuses on provide explicit instruction to prompting LLMs (e.g., Chain-of-Thought (CoT) and In-context Learning (ICL)).

CoT \cite{wei2022chain} aims to teach LLMs learning a manually well-designed reasoning pattern. Based on it, various technologies have been proposed, such as ToT \cite{yao2023tree}, which constructs a tree-based thinking process to perform deliberate decision-making process. CoN \cite{yu2024chain} generates sequential reading notes for each retrieved document to eliminate or reduce noisy information.

Furthermore, ICL has shown an advantage in the ability to solve complex tasks \cite{brown2020language}. For example, Structured Prompting \cite{hao2022structured} breaks the length limit and scales in-context learning to thousands of examples. WICL \cite{yang2023not} analyses the usefulness of examples in ICL, demonstrating the importance of example selection. Rishabh \textit{et al.} remove rationales and only provide domain-specific inputs , which also improve performance in ICL. IKE \cite{zheng2023can} utilizes ICL to teach LLMs to edit factual errors. Unlike previous work, we prefer to utilize partially relevant knowledge to awaken LLMs better utilizes the parametric knowledge rather than teach them a high-level reasoning pattern or simulate examples to analog inference.

\begin{figure*}[!ht]
    \centering
\includegraphics[scale=0.19]{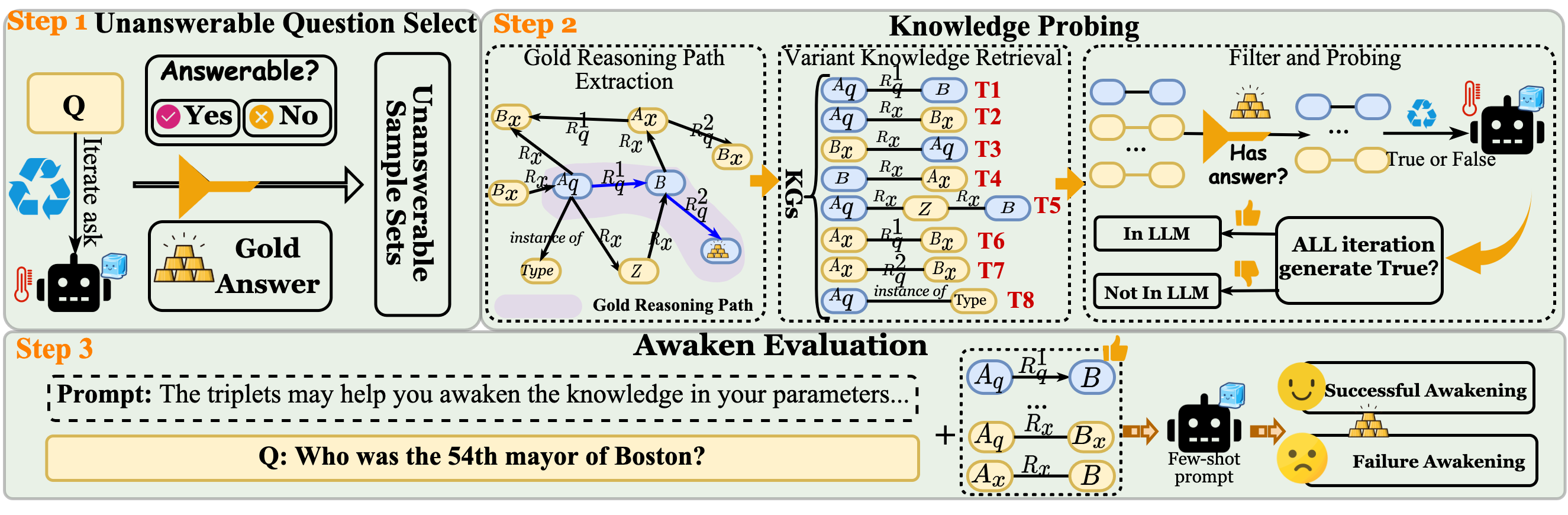}
    \caption{The process of Awaken LLMs. Step 1 selects the questions that can not be answered with the present LLM; Step 2 first extracts the gold reasoning path and then retrieves variant knowledge in a 2-hop subgraph of \(A_q\), which is dropped if it has the answer, the others are used to probe whether stored in LLMs via an iterative ``True of False'' QA. Step 3 utilizes that stored knowledge to evaluate the potential of awakening. \(A_x\),\(B_x\),\(Z\) and \(R_x\) are the arbitrary entity and relation in KG.}
    \Description{Illustration showing the analogy between human brain activation and large language model knowledge awakening.}
    \label{fig:framework}
\end{figure*}

\section{Task Formulation} 
\label{task formulation}
\textbf{Task1: Awakening LLMs} refers to utilizing incomplete explicit knowledge stored within the models to stimulate their internal knowledge for inference. Without losing generality, considering question answering (QA) as a case, given a question \(\mathcal{Q}\) and its gold answer \(\mathcal{A}\), if there is some partially relevant knowledge \(\mathcal{K}\) that do not contain \(\mathcal{A}\) and are embedded in an LLM \(\mathcal{M}\), such that:
\begin{equation}
     \mathcal{M}(\mathcal{Q}) \centernot\implies \mathcal{A} \; \text{and} \;
     \mathcal{M}(\mathcal{Q,K}) \implies \mathcal{A},
\end{equation}
then we say the LLM \(\mathcal{M}\) is awakened by \(\mathcal{K}\). 
The awakening process has two constraints: the \textit{first} is \(C1:\) \(\mathcal{K} \in {\mathcal{M}}\) which means that by the knowledge probing \cite{alivanistos2022prompting} (See section \ref{4.2}), \({\mathcal{M}}\) can infer \(\mathcal{K}\) that is some kind of fact or statement in the format of unstructured text or other forms like triple. The \textit{second} is \(C2:\) \(\mathcal{Q}\) cannot be answered by \(\mathcal{M}\) without any explicit knowledge.

\textbf{Task2: Unseen Entity KGQA} is to answer an NL question \(\mathcal{Q}\) with an entity from a given KG \(\mathcal{G}\), where NL represents the head entity \(e\) that cannot be matched in \(\mathcal{Q}\) and did not appear in \(\mathcal{G}\). This task can be defined as:
\begin{equation}
    \mathcal{F} :(\mathcal{Q,G}) \implies \mathcal{A}, 
\end{equation}
where the relation \(\mathcal{R}\) mentioned in \(\mathcal{Q}\) can also be matched in \(\mathcal{G}\) (\(\mathcal{R \in G}\)), \(\mathcal{G}\) is strictly incomplete due to \(e\) not matched in \(\mathcal{Q}\). In the era of LLMs, much work utilizes LLMs in the KGQA task, i.e., RoG \cite{luoreasoning} and ToG \cite{sunthink}, which retrieve triplets from a subgraph of \(G\). Different from \(\mathcal{G}\), \(e \in G\).

\begin{figure*}[h]
    \centering
\includegraphics[scale=0.18]{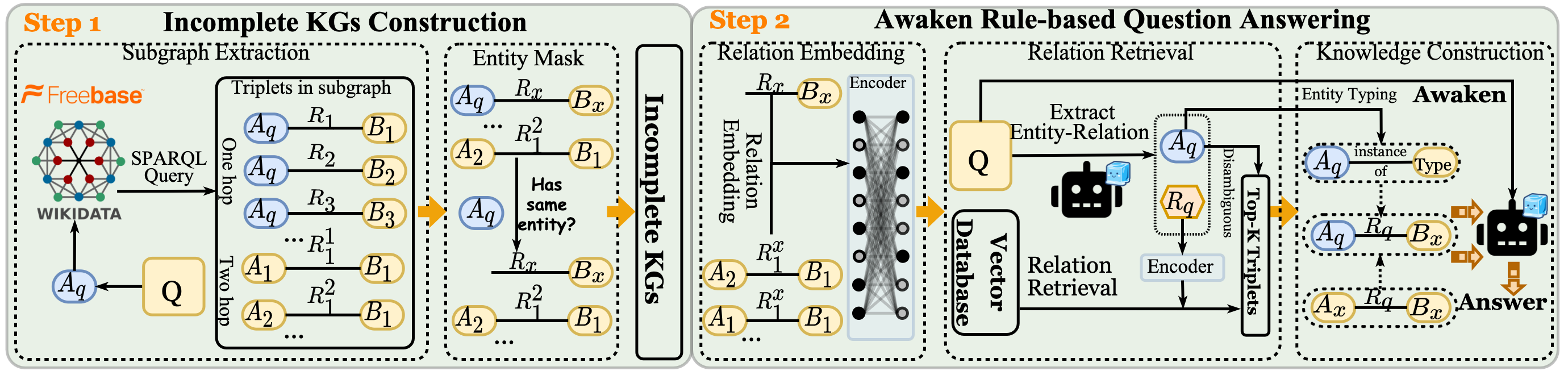}
    \caption{The process of Unseen Entity QA. We first retrieve the 2-hop subgraph and then mask the entity of triplet that contains the \(A_q\) to construct Incomplete KGs, which are transformed into vector database in Step 2. We then retrieve the top-k triplets according to the relation and entity and construct three type knowledge to awaken LLMs answer the question. \(R_x\) and \(R_x^y\) are the one-hop and two-hop relation in subgraph, respectively.}
    \Description{Illustration showing the analogy between human brain activation and large language model knowledge awakening.}
    \label{fig:framework2}
\end{figure*}

\section{Awakening LLMs with Partially Relevant Knowledge (Task1)}
This section mainly introduces two parts, including: (1) Theoretical analysis of LLMs awakening via the attention mechanism and the theory of Markov, (2) Benchmarking method to evaluate the performance of awakening through the partially relevant knowledge and its variants.

\subsection{Theoretical Analysis of LLMs Awakening}
\textbf{Attention mechanism.}The transformer architecture is the majority of choice for building LLMs. The core component is the attention mechanism that calculates the attention score with three variables: \(\mathbf{Q},\mathbf{K}\) and \(\mathbf{V}\).

We first define the representation of an initial input sequence as \(\mathbf{H}^{0} \in \mathbb{R}^{N\times d}\), for \(l\)-th (\(l=1,...,L\)) transformer, we define
\begin{equation}
    \mathbf{Q}^{(l)}=H^{(l-1)}\mathbf{W}_Q^{(l)}, \mathbf{K}^{(l)}=H^{(l-1)}\mathbf{W}_K^{(l)},  \mathbf{V}^{(l)}=H^{(l-1)}\mathbf{W}_V^{(l)},
\end{equation}
where \(\mathbf{W}_Q^{(l)}\),\(\mathbf{W}_K^{(l)}\),\(\mathbf{W}_V^{(l)} \in \mathbb{R}^{d \times d_{head}}\). \(N\) is the sequence length, \(d\) is the dimension of \(\mathbf{Q}^{(l)}, \; \mathbf{K}^{(l)} \; \mathbf{V}^{(l)}\). The calculation process of attention matrix can be described as:
\begin{equation}
    \mathbf{A}^{(l)}=\text{softmax}(\frac{\mathbf{Q}^{(l)}(\mathbf{K}^{(l)})^{\top }}{\sqrt{d_{head}}}) \in \mathbb{R}^{N\times N}.
    \label{eq2}
\end{equation}
\textbf{The theory of Markov in LLMs.} In the \(l\)-th layer of LLMs that is composed of \(L\) transformer layers, the attention matrix \(A^{(l)}\) can be interpreted as a context-conditioned Markov transition matrix. Although this transition is not temporal in the conventional sense, it satisfies the normalization and locality properties that characterize Markov transitions \cite{ildiz2024self}. The output of the \(l\)-th layer is calculated as \(\mathbf{H}^{(l)}=\mathbf{A}^{(l)}\mathbf{V}^{(l)}\). As our analysis emphasizes structural connectivity across layers instead of specific numerical operations, we abstract components such as residual links and multi-head attention aggregation as identity matrices \(\mathbf{I}\). The computational relationship between successive layers is given by \(\mathbf{H}^{(l)}=\mathbf{A}^{(l-1)}\mathbf{V}^{(l-1)}=\mathbf{A}^{(l-1)}(\mathbf{H}^{(l-1)}\mathbf{W}_{V}^{(l-1)})\).

Based on this interpretation, we abstract the attention-driven token interaction at each layer as a Markovian transition over the input sequence.  Due to the attention patterns vary across layers, \(\mathbf{A}^{(1)} \ne \mathbf{A}^{(2)} \ne ...\ne \mathbf{A}^{(L)}\). The entire propagation process can be regarded as a non-homogeneous Markov chain, where each layer acts as a distinct time step with its own transition operator. Consequently, the final representation at the top layer \(H^{(L)}\) is obtained via a composition of layer-wise Markov transitions:
\begin{equation}
    \label{eq5}
    \mathbf{H}^{(L)}=( \textstyle \prod_{l=1}^{L}\mathbf{A}^{(l)})\mathbf{H}^{(0)}=\mathbf{A}^{(L)}\mathbf{A}^{(L-1)}...\mathbf{A}^{(1)}\mathbf{H}^{(0)}.
\end{equation}

We refer to this hierarchical formulation as a layer-wise Markov propagation model for attention in deep transformer architectures (e.g., LLMs). In light of this, we define \(\mathbf{H}^{(0)}\) is the hidden states of \(\mathcal{Q \;\text{and} \;K}\), (\(\textstyle \prod_{l=1}^{L}\mathbf{A}^{(l)}), \mathbf{H}^{(0)}\) is the structured latent activation of \(\mathcal{K'}\) in LLM, \(\mathbf{H}^{(L)}\) is the final output hidden states of the LLM, which encode the model's answer representation \(\mathcal{A}\).

In the definition of Task1, the condition \(\mathcal{K} \in M\) implies that \(\mathbf{H}^{(0)}\) contains at least one non-zero element, i.e., \(\exists i,j \in \{1,...,n\}, \mathbf{H}_{i,j}^{(0)} \neq 0\). Given that the token \(\mathcal{A}\) appears later in the sequence than the tokens \(\mathcal{Q}\) and \(\mathcal{K}\), the causal mask in LLMs ensures \(A^{(l)} > 0\), which guarantees the \textit{structural feasibility} of an attention path from \(\mathcal{Q,K}\) to \(\mathcal{A}\). Consider \(\mathbf{H}_{i,j}^{(0)} \neq 0\) and \(A^{(l)} > 0\), \(\exists i,j,\; \mathbf{H}_{i,j}^{(L)} > 0\), which indicates the existence and \textit{computability of an effective attention path} from \(\mathcal{Q}\) and \(\mathcal{K}\) through structured latent activation knowledge \(\mathcal{K'}\) to \(\mathcal{A}\). Based on this observation, our objective is to identify the optimal \(\mathcal{K}\) that maximizes the likelihood of generating a correct answer in LLMs under specific conditions.

\begin{table*}[]
\centering
\caption{8 types of awakening knowledge for question ``Team owner \colorbox{blue!15}{Tom Gores} won the Championship in what year?''}
\label{tab:type}
\resizebox{\textwidth}{!}{%
\begin{tabular}{lll}
\toprule
\multicolumn{1}{c}{Knowledge Type} & \multicolumn{1}{c}{Definition} & \multicolumn{1}{c}{Examples} \\ \midrule
\(A_q\)-\(R_q^1\)-\(B\) (\textbf{T1}) & \colorbox{blue!15}{Start entity} with first hop \colorbox{yellow!15}{gold outgoing relation} & \colorbox{blue!15}{Tom Gores}, \colorbox{yellow!15}{owne\_s}, \colorbox{cyan!15}{Detroit Pistons} \\
\(A_q\)-\(R_x\)-\(B_x\) (\textbf{T2}) & \colorbox{blue!15}{Start entity} with non-gold outgoing relation & \colorbox{blue!15}{Tom Gores}, Organizations founded, Platinum Equity \\
\(B_x\)-\(R_x\)-\(A_q\) (\textbf{T3}) & \colorbox{blue!15}{Start entity} with non-gold incoming relation & Financier, People With This Profession, \colorbox{blue!15}{Tom Gores} \\
\(A_q\)-\(\text{instance-of}\) (\textbf{T4}) & Type information of \colorbox{blue!15}{start entity} & \colorbox{blue!15}{Tom Gores}, instance-of, Sports Team Owner \\
\(B\)-\(R_x\)-\(A_x\) (\textbf{T5}) & \colorbox{cyan!15}{Intermediate entity} with non-gold outgoing relaiton & \colorbox{cyan!15}{Detroit Pistons}, Team Mascot, Hooper \\
\(A_q\)-\(R_x\)-\(Z\)-\(R_x\)-\(B\) (\textbf{T6}) & \colorbox{cyan!15}{Intermediate entity} with non-gold incoming relation & \colorbox{blue!15}{Tom Gores}, type,Agent; Agent, Instance, \colorbox{cyan!15}{Detroit Pistons} \\
\(A_x\)-\(R_q^1\)-\(B_x\) (\textbf{T7}) & Arbitrary entities with \colorbox{yellow!15}{gold first hop relation} & Greenville Drive \colorbox{yellow!15}{owne\_s} RB3, LLC \\
\(A_x\)-\(R_q^2\)-\(B_x\) (\textbf{T8}) & Arbitrary entities with \colorbox{magenta!15}{gold second hop relation} & US men's soccer team, \colorbox{magenta!15}{championships}, 2005 CONCACAF Gold Cup \\ \bottomrule
\end{tabular}%
}
\end{table*}

\subsection{Benchmarking Method}
\label{Benchmarking}
As shown in Figure \ref{fig:framework}, the evaluation method in Task1 has three steps: \textbf{Step 1: Unanswerable question selection} selects the samples from 2Wiki and CWQ to satisfy \(C2\); \textbf{Step 2: Knowledge Probing} extracts the gold reasoning path and its variant knowledge from the KG for those unanswerable questions, and selects knowledge satisfying \(C1\) through answer filtering and probing; \textbf{Step 3: Awaken Evaluation} evaluates the awakening phenomenon in LLMs using prior knowledge. Details of the process are as follows:
\label{4.2}

\textbf{Step 1: Unanswerable question selection.} As described fundamental conditions in Section \ref{task formulation}, we perform a data filtering process to find \(\mathcal{Q}\) that satisfies \(C2\): prompting LLMs to answer questions without additional knowledge to divide an existing dataset into an answerable and unanswerable subset. To ensure that the given question is definitively unanswerable, we randomly set a different temperature \(T_i\) which is in \(\{T_1,T_2,T_3\}\), and ranges from 0.1 to 1.0 for LLMs to answer the question three times. If all responses cannot be correctly answered, this question is defined as unanswerable for the present LLM, which formalized as:
\begin{equation}
    \mathbb{Q}_{\mathrm{un}} = \{\mathcal{Q} \in \mathbb{Q}| \; \forall T_i \in \mathcal{T}, \mathrm{Correct}(\mathcal{M}_{T_i}(\mathcal{Q}),\mathcal{A})=0\},
\end{equation}
where \(\mathbb{Q}_{\mathrm{un}}\) represents a subset of the total questions \(\mathbb{Q}\) that are unanswerable, \(\mathrm{Correct}(\cdot,\cdot)\) is a function that calculates the correctness between the LLMs' response \(\mathcal{M}_T(\mathcal{Q})\) and the gold answer \(\mathcal{A}\) via \(\mathrm{ACC_R}\) and returns 1 for correct and 0 for error.

\textbf{Step 2: Knowledge Probing.}
This step has three sub-stages to provide awakening knowledge for evaluation:

\textit{Stage 1: Gold Reasoning Path Extraction:} This stage extracts the gold reasoning path from the KG for those unanswerable questions \(\mathbb{Q}_{\mathrm{un}}\). The path has two type of entities: the start and intermediate entities (\(A_q\) and \(B\)) can be used to evaluate, and five types of relations: gold incoming and outgoing, non-gold incoming and outgoing, and entity type relation, which are denoted as \(R_q\), \(R_x\) and \(\text{instance-of}\), respectively. We extract \(A_q\)-\(R_q^1\)-\(B\) (\textbf{T1}) and \(R_q^2\) according to the information provided in the datasets, \(R_q^1\) and \(R_q^2\) represent the one-hop and two-hop gold reasoning relations.

\textit{Stage 2: Variant Knowledge Retrieval:} This stage retrieves some variant knowledge from KGs based on the gold reasoning path. As shown in Table \ref{tab:type}, we derive variants in the forms of \(A_q\)-\(R_x\)-\(B_x\) (\textbf{T2}), \(B_x\)-\(R_x\)-\(A_q\) (\textbf{T3}), \(A_q\)-\(\text{instance-of}\) (\textbf{T4}), \(B\)-\(R_x\)-\(A_x\) (\textbf{T5}), \(A_q\)-\(R_x\)-\(Z\)-\(R_x\)-\(B\) (\textbf{T6}), \(A_x\)-\(R_q^1\)-\(B_x\) (\textbf{T7}) and \(A_x\)-\(R_q^2\)-\(B_x\) (\textbf{T8}). The retrieval process first maps \(A_q\) into a Wikidata entity (whose ID starts with Q) utilizing the wikimapper\footnote{https://github.com/jcklie/wikimapper}. Subsequently, the subgraph\footnote{https://query.wikidata.org/} is queried to extract the triplets that manifest the same pattern as the variant knowledge. It is noteworthy that all variant awakening knowledge was only searched in the two-hop subgraph of \(A_q\) and \(B\), because the triplet too far from them is useless; the latency and the number of triplets cost too much resource. The knowledge that has the same pattern may exist multiple triplets, a reranking model (bge-m3-reranker \cite{chen2024bge}) is used to rank these knowledge according to \(A_q\)-\(R_q\)-\(B\), and the top-1 is selected as the final variant knowledge. 

\textit{Stage 3: Filter and Probing:} This stage filters the triplets that contain the gold answer and probes whether the knowledge is embedded in LLMs to satisfy \(C2\). In detail, for every knowledge acquired in the earlier stages, it is filtered when the gold answer is contained. The others are used to probe with an ``True or False'' QA task. This process iterates three times, if all responses of LLMs are ``True'', the corresponding knowledge is confirmed to exist in LLMs. This process can be defined as:
\begin{equation}
    \mathrm{Probe(\mathcal{K})=}\begin{cases}
 \mathrm{True} & \text{ if } \forall T_i \in \mathcal{T}, \mathcal{M}_T(P(\mathcal{K} ))=\text{``}\mathrm{True}\text{''} \\
 \mathrm{False} & \mathrm{otherwise}.
\end{cases}
\end{equation}

\begin{table*}[t]
\small
\centering
\caption{The awaken experimental results on two datasets across 5 Instruct version LLMs. ``F'' indicates the number of samples are used to test, which after filter in section \ref{4.2}. \colorbox{red!10}{\textcolor{red!10}{\rule{1em}{0.8ex}}} represents No RAG results, \colorbox{green!10}{\textcolor{green!10}{\rule{1em}{0.7ex}}} indicates the awaken results. The best two results are highlighted by \colorbox{orange!60}{1st} and \colorbox{myyellow}{2nd}. More evaluation results of awakening knowledge type can be found in Appendix \ref{app:res}.} 
\label{tab:same_question}
\setlength{\tabcolsep}{0.13cm}{%
\begin{tabular}{ccclcclcclcclccl}
\toprule
\multicolumn{1}{c}{\multirow{3}{*}{\textbf{Knowledge type}}} & \multicolumn{3}{c}{\textbf{Llama3.1-8B}} & \multicolumn{3}{c}{\textbf{Llama3.1-70B}} & \multicolumn{3}{c}{\textbf{Qwen2.5-7B}} & \multicolumn{3}{c}{\textbf{Qwen2.5-32B}} & \multicolumn{3}{c}{\textbf{Qwen2.5-72B}} \rule{0pt}{2ex}\\ \cline{2-16}
\multicolumn{1}{c}{} & F & Hits@1 & Hits@10 & F & Hits@1 & Hits@10 & F & Hits@1 & Hits@10 & F & Hits@1 & Hits@10 & F & Hits@1 & Hits@10 \rule{0pt}{2ex}\\ \cline{2-16} 
 & \multicolumn{15}{c}{\textbf{2Wiki}} \rule{0pt}{2ex}\\ \cline{2-16}
 \rowcolor{red!10}
No RAG & 815 & 0.0 & 13.3 & 693 & 0.0 & 16.2 & 289 & 0.0 & 4.8 & 323 & 0.0 & 4.3 & 306 & 0.0 & 3.9 \rule{0pt}{2ex}\\
\rowcolor{green!10}
\textbf{\(A_q\)-\(R_q^1\)-\(B\)} & 815 & 3.7 & \cellcolor{myyellow}{\centering 15.5} & 693 & \cellcolor{myyellow}{\centering 5.9} & \cellcolor{myyellow}{\centering 17.6} & 289 & \cellcolor{myyellow}{\centering 2.8} & 5.2 & 323 & 5.3 & 13.0 & 306 & 4.9 & 11.1 \\
\rowcolor{green!10}
\textbf{\(A_q\)-\(R_x\)-\(B_x\)} & 815 & \cellcolor{myyellow}{\centering 4.4} & 14.8 & 693 & 4.6 & 15.6 & 289 & 2.4 & 4.5 & 323 & 4.3 & 10.2 & 306 & 6.9 & 11.1 \\
\rowcolor{green!10}
\textbf{\(B\)-\(R_x\)-\(A_x\)} & 815 & 3.1 & \cellcolor{orange!60}{\centering 16.4} & 693 & 5.8 & 17.5 & 289 & 1.4 & \cellcolor{myyellow}{\centering 5.5} & 323 & 5.6 & \cellcolor{myyellow}{\centering 13.6} & 306 & 6.2 & \cellcolor{orange!60}{\centering 13.7} \\
\rowcolor{green!10}
\textbf{\(A_x\)-\(R_q^1\)-\(B_x\)} & 815 & 3.3 & 13.5 & 693 & \cellcolor{orange!60}{\centering 6.1} & 15.7 & 289 & \cellcolor{orange!60}{\centering 4.2} & \cellcolor{orange!60}{\centering 7.3} & 323 & \cellcolor{myyellow}{\centering 6.2} & \cellcolor{orange!60}{\centering 13.9} & 306 & \cellcolor{myyellow}{\centering 6.5} & \cellcolor{myyellow}{\centering 11.8} \\
\rowcolor{green!10}
\textbf{\(A_q\)-\(R_x\)-\(Z\)-\(R_x\)-\(B\)} & 815 & 4.0 & 12.6 & 693 & 5.8 & 15.9 & 289 & 1.7 & 4.5 & 323 & \cellcolor{orange!60}{\centering 6.8} & 11.1 & 306 & \cellcolor{orange!60}{\centering 8.8} & \cellcolor{orange!60}{\centering 13.7} \\
\rowcolor{green!10}
\(A_q\)-\(\mathrm{instance}\)-\(\text{of}\) & 815 & \cellcolor{orange!60}{\centering 4.7} & 12.9 & 693 & 4.6 & \cellcolor{orange!60}{\centering 19.8} & 289 & 1.0 & 3.1 & 323 & 3.4 & 8.4 & 306 & 4.6 & 10.5 \\ \cline{2-16} 
 & \multicolumn{15}{c}{\textbf{CWQ}} \rule{0pt}{2ex}\\ \cline{2-16} 
 \rowcolor{red!10}
No RAG & 446 & 0.0 & 30.7 & 375 & 0.0 & 19.5 & 353 & 0.0 & 12.5 & 256 & 0.0 & 9.0 & 270 & 0.0 & 8.9\rule{0pt}{2ex} \\
\rowcolor{green!10}
\textbf{\(A_q\)-\(R_q^1\)-\(B\)} & 446 & \cellcolor{orange!60}{\centering 39.2} & 56.3 & 375 & 33.3 & 45.1 & 353 & 25.2 & \cellcolor{myyellow}{\centering 38.5} & 256 & \cellcolor{myyellow}{\centering 37.1} & 45.7 & 270 & 25.9 & 33.3 \\
\rowcolor{green!10}
\textbf{\(A_q\)-\(R_x\)-\(B_x\)} & 446 & 34.1 & 54.9 & 375 & 32.8 & 45.3 & 353 & 26.1 & 38.2 & 256 & \cellcolor{myyellow}{\centering 37.1} & 46.1 & 270 & 28.9 & \cellcolor{myyellow}{\centering 35.2} \\
\rowcolor{green!10}
\textbf{\(B\)-\(R_x\)-\(A_x\)} & 446 & \cellcolor{myyellow}{\centering 37.0} & \cellcolor{orange!60}{\centering 58.7} & 375 & \cellcolor{orange!60}{\centering 34.4} & \cellcolor{orange!60}{\centering 46.1} & 353 & \cellcolor{myyellow}{\centering 27.5} & 37.7 & 256 & \cellcolor{orange!60}{\centering 38.7} & \cellcolor{orange!60}{\centering 50.4} & 270 & \cellcolor{myyellow}{\centering 29.6} & \cellcolor{orange!60}{\centering 35.6} \\
\rowcolor{green!10}
\textbf{\(A_x\)-\(R_q^1\)-\(B_x\)} & 446 & 16.4 & 39.0 & 375 & 12.5 & 24.3 & 353 & 7.6 & 16.4 & 256 & 16.4 & 32.4 & 270 & 17.8 & 28.5 \\
\rowcolor{green!10}
\textbf{\(A_q\)-\(R_x\)-\(Z\)-\(R_x\)-\(B\)} & 446 & 34.3 & 54.3 & 375 & \cellcolor{myyellow}{\centering 33.9} & \cellcolor{myyellow}{\centering 45.9} & 353 & 20.4 & 30.9 & 256 & 32.4 & 44.5 & 270 & \cellcolor{orange!60}{\centering 30.4} & 34.1 \\
\rowcolor{green!10}
\(A_q\)-\(\mathrm{instance}\)-\(\text{of}\) & 446 & 33.4 & \cellcolor{myyellow}{\centering 58.3} & 375 & 33.1 & 45.3 & 353 & \cellcolor{orange!60}{\centering 28.6} & \cellcolor{orange!60}{\centering 41.1} & 256 & 35.5 & \cellcolor{myyellow}{\centering 48.8} & 270 & 24.8 & 34.1 \\ \bottomrule
\end{tabular}%
}
\end{table*}

\textbf{Step 3: Awaken Evaluation.} This process utilizes six types of retrieved partial relevant knowledge \(\mathcal{K}\)—already embedded in LLMs—to evaluate whether they can help LLMs answer questions \(\mathcal{Q}\) that were initially unanswerable. If LLMs correctly answer the question after injecting the knowledge, which demonstrates that LLMs are successfully awakened, otherwise not. The prompt template can be found in Appendix \ref{app:prompt}.

\section{Unseen Entity QA (Task2)}
\label{5.1}
This section describes the incomplete KG construction that retrieves the two-hop subgraph of the entity and an awaken rule-based method that utilizes the relation appearing in question to retrieve and construct awakening knowledge for Unseen Entity QA task.

\subsection{Incomplete KGs Construction}
As shown in Figure \ref{fig:framework2} left, this stage introduces the procedure that leverages the existing KGs to formulate incomplete KGs for Unseen Entity QA, which has two sub processes:  \textbf{Subgraph Extraction} and \textbf{Entity Mask}. The previous extracts a two-hop subgraph of the entity \(A_q\) mentioned in the question \(\mathcal{Q}\) via ``Qid'', the one- and two-hop triplets are formalized as \(A_q\)-\(R_x\)-\(B_x\) and \(B_x\)-\(R_x^y\)-\(C_x\), respectively. The extract process is used for every question in the dataset, and all triplets are composed of initial KG. \textbf{Entity Mask} then masks the entity present in both the triplets and the questions to construct the definitive incomplete KG \(\mathcal{G}\), thereby emulating the Unseen entity QA wherein the entity cannot be associated with KG.

\subsection{Awaken Rule-based Method}
In this section, an awaken rule-based QA method (Figure \ref{fig:framework} (right)) was proposed for the Unseen Entity QA task. Specifically, this method contains three stages: 1) \textbf{Relation Embedding} will embed the relation \(\mathcal{R}\) as the vector \(E(\mathcal{R})\) of whole triple for retrieving knowledge; 2) \textbf{Relation Retrieval} employs the relation \(R_q\) to retrieve relevant knowledge; 3) \textbf{Knowledge Construction} formulates specific knowledge according to the awaken rule, which was used to answer the question.

\textbf{Relation Embedding.} Due to the incompleteness of the KG, the entity linking based KG-RAG methods \cite{wang-etal-2021-retrieval, he2024g, sunthink, luoreasoning} can not retrieve on the subgraph of question, we retrieve the awakening knowledge from the total incomplete KG. Since \(A_q \notin \mathcal{G}\) and \(R_q \in \mathcal{G}\), the retrieval process can only be conducted based on the relation provided in the question and triplet, so that we embed the relation \(R_x\) to represent the entire triplet with a sentence embedding language model \(E\) \footnote{https://huggingface.co/sentence-transformers/all-MiniLM-L6-v2}. This process can be defined as \(\mathbf{e}_\mathcal{K}=E(R_x)\).

\textbf{Relation Retrieval.} Given that \(\mathcal{G}\) comprises millions of triples, we employ an offline indexing strategy to ensure retrieval efficiency. Note that, we utilize FAISS to indexing and similarity calculation, since it provides many strategies to decrease the cost of retrieval time in billions of vectors. \(R_x \in \mathcal{G}\) has the limited types, the Inverted File Index with Flat Quantization is used as an approximate Nearest-Neighbor method to search the relevant vector. The index construct process can be formalized as \(\mathcal{I}=\mathrm{IndexIVFFlat}(\mathbf{e}_\mathcal{K\in G} )\).

For \(\mathcal{Q}\), we first extract the entity \(A_q\) and its relation \(R_q^1\) that is used to retrieve top-k triplets via the same embedding model in Relation Embedding. \(A_q\) is used to disambiguous that those triplets have the same relation with \(R_q^1\) with LLMs, outputting the most useful triplets \(A_x\)-\(R_q^1\)-\(B_x\). Due to the fact that the entity of relation \(R_q^2\) is unclear, making it difficult to extract and use it exactly in the retrieval. Throughout the remainder of this paper, we adopt \(R_q^1\) as the sole retrieval signal, denoted concisely as \(R_q\) in the Unseen Entity KGQA task. This retrieval process can be defined as \(\mathcal{K}=\mathcal{I}(E(R_q))\). Details are shown in Appendix \ref{app:exp}.

\textbf{Knowledge Construction.} Since \(A_q \notin \mathcal{G}\) and \(B\) are not available directly, T1-T3 and T6 are not constructed to answer the question. So we roughly construct the target knowledge \(A_q\)-\(R_q\)-\(B\) by combining the retrieved \(A_x\)-\(R_q\)-\(B_x\) or \(R_q\)-\(B_x\) with \(A_q\), which is denoted as \(A_q\)-\(R_q\)-\(B_x\). Furthermore, \(A_q\)-\(\mathrm{instance}\)-\(\text{of}\) can also be used to awaken, we leverage the schema information from YAGO 4.5 by extracting the 40 top-level types (e.g., human, country). Then, using \(\mathcal{Q}\) as context, we prompt LLMs to infer the most likely type of \(A_q\). Finally, we obtain three types of knowledge: \(A_x\)-\(R_q\)-\(B_x\) or \(R_q\)-\(B_x\) (referring to \textbf{T7}), \(A_q\)-\(R_q\)-\(B_x\) (denoted \textbf{T1*}) and \(A_q\)-\(\mathrm{instance}\)-\(\text{of}\) (\textbf{T4}), which are individually or collectively inputted with the question to awaken the LLMs. Due to its poor awakening performance in Figure \ref{fig:heatmap}, \(B_x\)-\(R_x\)-\(A_q\) (T3) is excluded.

\section{Evaluation and Results}
In our evaluation, we aim to answer the following research questions: 
\textbf{RQ1}: How do different types of knowledge affect the performance of knowledge awakening?
\textbf{RQ2}: What are the factors that determine the performance of knowledge awakening?
\textbf{RQ3}: What is the performance of the awakening mechanism under an incomplete KGs retrieval?

\subsection{Experiments Settings}
This section mainly introduces the datasets, metrics used and baseline method in Unseen Entity QA task.

\textbf{Datasets.} We perform extensive experiments on three Question Answering datasets, including: 2WikiMultiHopQA (\textbf{2Wiki}), ComplexWebQuestions (\textbf{CWQ}) and \textbf{Mintaka} \cite{sen2022mintaka}. For awaken LLMs, we use 2Wiki and CWQ because they provide the gold reasoning path from KGs and for Unseen Entity QA, we use all of the dataset for evaluate. The detail of datasets can be found in Appendix \ref{app:exp}.

\textbf{Metrics.}
We assess the awaken ability through several metrics: Hits@1 and Hits@10. The hit of the answer is used \(\mathrm{ACC_R}\) to calculate, which determines whether the golden answer is contained within the response of LLMs \cite{sun2025rearter, song2025r1}.

\textbf{Baselines.} The proposed method compared with serval baselines, which contains a \textbf{No RAG} and two RAG methods, including: \textbf{DiFaR} \cite{baek2023direct} and integrating Question Decomposition \cite{ye2023large} with DiFaR (\textbf{QD-DiFaR}). We consider two settings for DiFaR and QD-DiFaR baselines: (1) Embedding the question as a query to retrieve; (2) Embedding the entity mentioned in the question as a query. Details about baselines can be found in Appendix \ref{app:exp}.

\begin{figure}[t]
    \centering
    \includegraphics[scale=0.4]{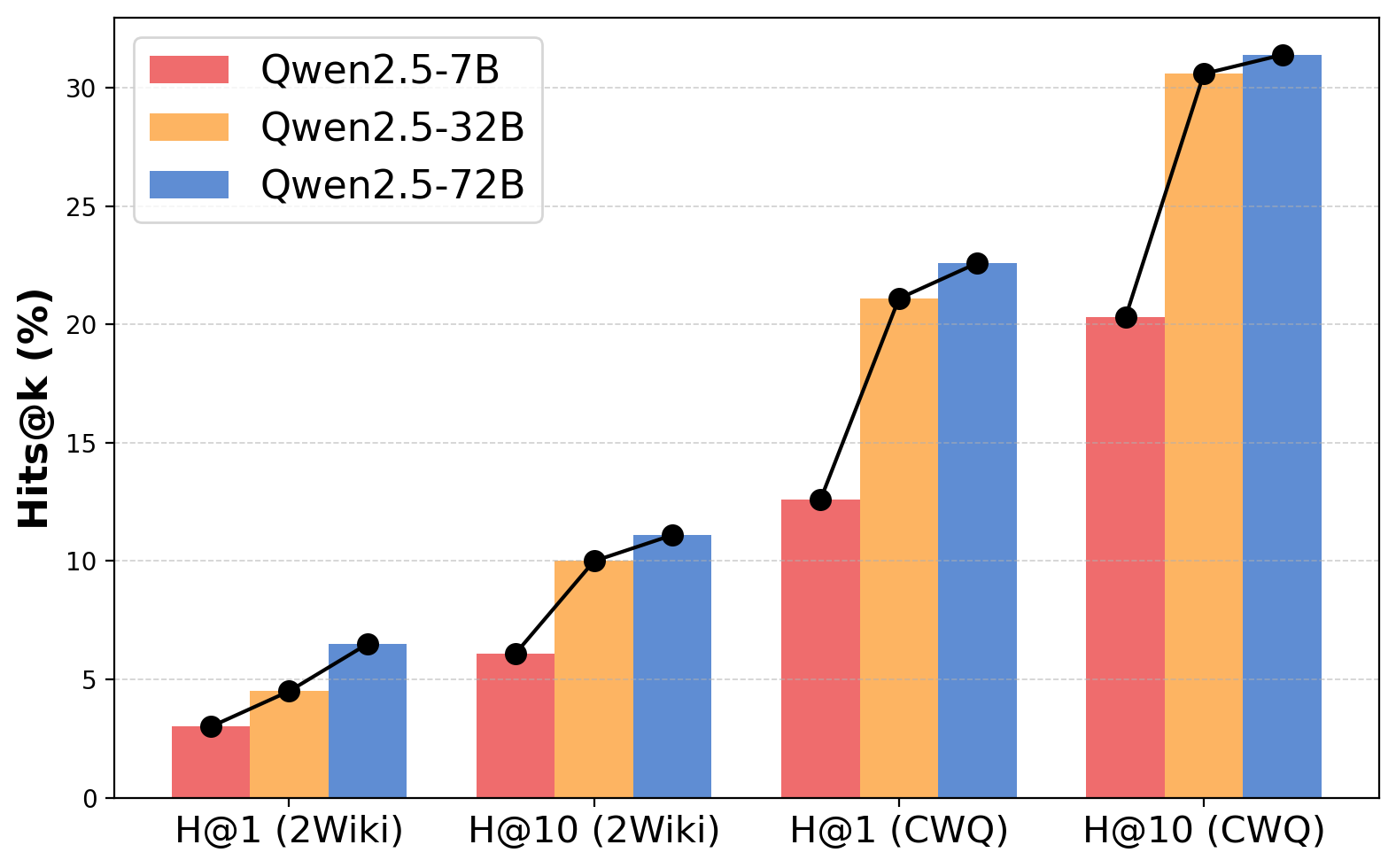}
    \caption{Impact of model size on CWQ. The x-axis represents metric type (Hits@k refer to H@k) and dataset, and the y-axis shows the Hits@k score. See Appendix \ref{size} for full results.}
    \label{different_llm}
\end{figure}

\begin{figure}[t]
    \centering
    \includegraphics[scale=0.33]{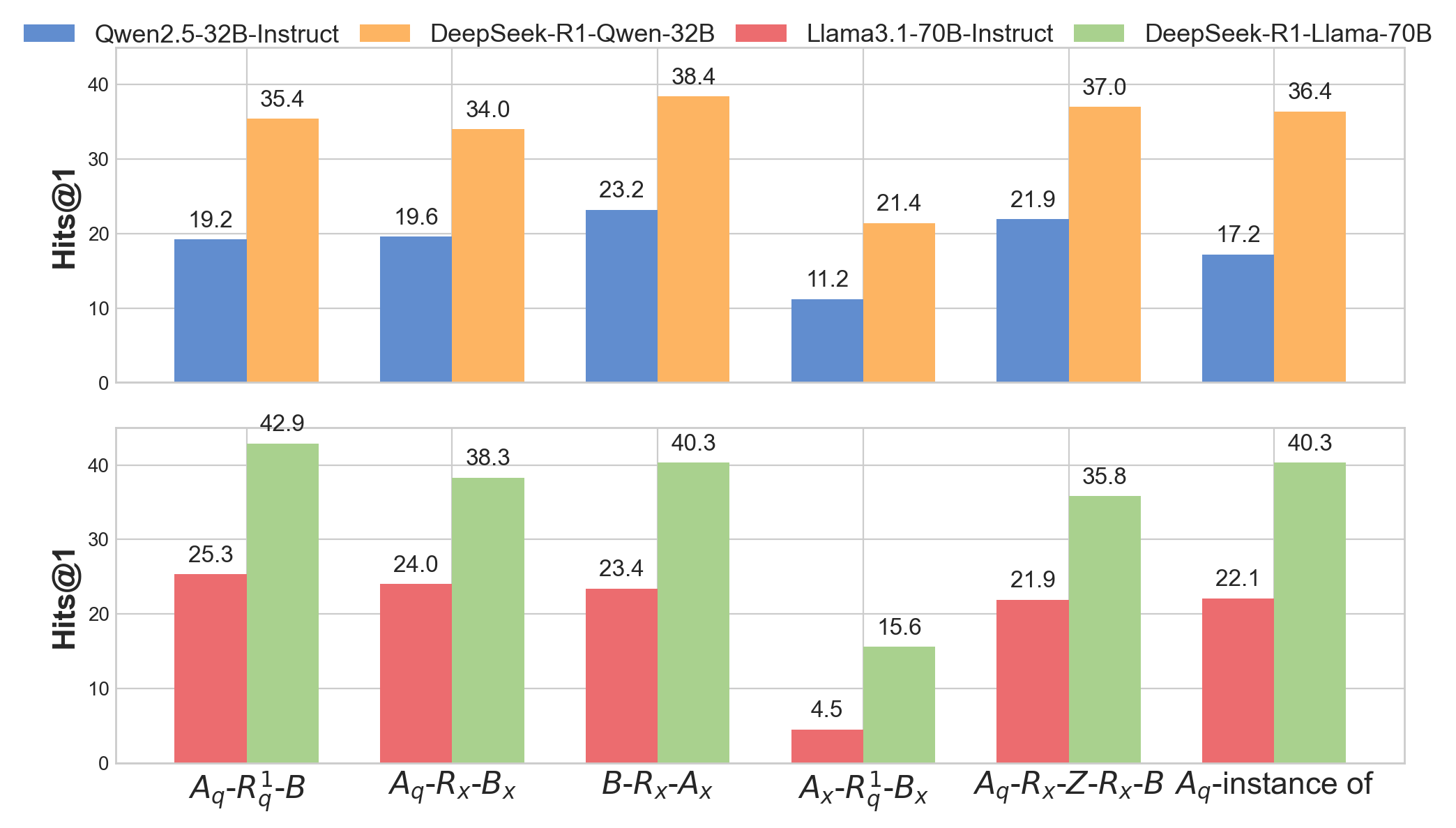}
    \caption{Comparison of performance between Instruct-version models and R1-style reasoning models. x-axis is different awakening knowledge type, y-axis is Hits@1 score.}
    \label{fig:instruct_r1}
\end{figure}

\subsection{RQ1: How do Different types of Knowledge Affect the Knowledge Awakening?}

\begin{table*}[t]
\small
\centering
\caption{The performance in Unseen Entity QA. (T1*) represents only use \(A_q\)-\(R_q\)-\(B_x\) to awaken. (T4 and T1*) indicate that utilizing multiple types knowledge at same times. H@1 and H@10 are the Hits@1 and Hits@10 score, each accordingly. (entity) means using entity as quety. The best two results are highlighted by \colorbox{orange!60}{1st} and \colorbox{myyellow}{2nd}. GraphRAG methods are not compatible with KG inputs, as they require documents, and thus are excluded from comparison. KG-RAG requires subgraphs of the target entity, which is infeasible in the Unseen Entity QA setting.}
\label{tab:qa}
\setlength{\tabcolsep}{0.068cm}{%
\begin{tabular}{lcccccccccccccccccc}
\hline
\multicolumn{1}{c}{\multirow{3}{*}{Method}} & \multicolumn{2}{c}{Mintaka} & \multicolumn{2}{c}{2Wiki} & \multicolumn{2}{c}{CWQ} & \multicolumn{2}{c}{Mintaka} & \multicolumn{2}{c}{2Wiki} & \multicolumn{2}{c}{CWQ} & \multicolumn{2}{c}{Mintaka} & \multicolumn{2}{c}{2Wiki} & \multicolumn{2}{c}{CWQ} \\ \cline{2-19} 
\multicolumn{1}{c}{} & \multicolumn{6}{c}{Llama3.1-8B-Instruct} & \multicolumn{6}{c}{Qwen2.5-32B-Instruct} & \multicolumn{6}{c}{Llama3.1-70B-Instruct} \\ \cline{2-19} 
\multicolumn{1}{c}{} & H@1 & H@10 & H@1 & H@10 & H@1 & H@10 & H@1 & H@10 & H@1 & H@10 & H@1 & H@10 & H@1 & H@10 & H@1 & H@10 & H@1 & H@10 \\ \cline{2-19} 
\rowcolor{red!10}
No RAG & \colorbox{myyellow}{51.3} & 64.0 & 32.8 & 47.5 & 37.9 & 50.1 & \colorbox{myyellow}{55.9} & 61.0 & 32.7 & 37.9 & 43.6 & 43.6 &  \colorbox{myyellow}{66.1} & 75.0 & 39.5 & 58.1 & 51.0 & 63.2 \\
\rowcolor{red!10}
DiFaR (question) & 25.9 & 57.0 & 38.4 & 48.2 & 31.3 & 55.8 & 32.5 & 45.0 & 37.2 & 38.9 & 26.6 & 38.3 & 41.1 & 62.1 & 18.7 & 42.2 & 32.0 & 50.8 \\
\rowcolor{red!10}
QD+DiFaR (question) & 18.2 & 44.6 & 40.0 & 52.6 & 23.5 & 47.1 & 35.8 & 48.5 & 37.8 & 41.0 & 30.7 & 41.4 & 48.8 & 66.1 & 21.6 & 46.5 & 39.2 & 54.0 \\
\rowcolor{red!10}
DiFaR (entity) & 24.9 & 62.7 & 37.7 & 48.0 & 22.8 & 56.3 & 23.9 & 39.0 & 36.5 & 38.4 & 20.6 & 31.6 & 34.3 & 59.3 & 18.0 & 41.4 & 27.9 & 48.7 \\
\rowcolor{red!10}
QD+DiFaR (entity)& 27.8 & 60.8 & 39.2 & 51.9 & 30.1 & 56.1 & 27.9 & 44.4 & 37.6 & 40.2 & 20.7 & 33.9 & 43.4 & 65.5 & 22.5 & 45.0 & 32.3 & 50.8 \\
 \rowcolor{green!10}
\textbf{Ours (T7)} & 42.2 & \colorbox{myyellow}{68.2} & 41.2 & \colorbox{myyellow}{53.2} & \colorbox{orange!60}{40.0} & \colorbox{orange!60}{61.3} & 51.6 & 64.7 & 39.7 & 43.4 &  \colorbox{myyellow}{48.5} &  \colorbox{myyellow}{58.9} & 61.0 & 73.8 &  \colorbox{myyellow}{46.6} & 58.0 & \colorbox{orange!60}{54.0} & \colorbox{orange!60}{66.8} \\
 \rowcolor{green!10}
\textbf{Ours (T1*)} & 43.4 & 68.0 & \colorbox{myyellow}{41.9} & 52.3 & 36.8 & 58.1 & 53.9 & \colorbox{myyellow}{65.9} & 39.8 & 45.5 & 44.8 & 55.3 & 63.9 & 75.8 & 46.2 & 59.1 & 48.7 & 61.6 \\
 \rowcolor{green!10}
\textbf{Ours (T4)} & \colorbox{orange!60}{53.9} & \colorbox{orange!60}{75.2} & \colorbox{orange!60}{42.7} & \colorbox{orange!60}{53.9} & \colorbox{myyellow}{38.6} & \colorbox{myyellow}{59.0} & \colorbox{orange!60}{56.4} & \colorbox{orange!60}{66.6} & 40.4 & \colorbox{myyellow}{45.7} & 43.5 & 54.6 & 63.0 &  \colorbox{myyellow}{77.5} & \colorbox{orange!60}{48.8} & \colorbox{orange!60}{63.3} & 48.2 & 64.2 \\
 \rowcolor{green!10}
\textbf{Ours (T7 and T1*)} & 38.9 & 63.7 & 40.9 & 50.5 & 37.4 & 56.6 & 50.6& 63.1 & \colorbox{orange!60}{41.3} & \colorbox{orange!60}{46.1} & \colorbox{orange!60}{49.6} & \colorbox{orange!60}{59.1} & \colorbox{orange!60}{71.9} & \colorbox{orange!60}{83.5} & 45.6 & 57.5 & 53.4 &  \colorbox{myyellow}{65.1} \\
 \rowcolor{green!10}
\textbf{Ours (T7, T1* and T4)} & 37.0 & 61.4 & 40.8 & 51.0 & 36.5 & 55.5 & 50.3 & 63.6 & \colorbox{myyellow}{40.6} & 44.7 & 46.9 & 57.5 & 59.4 & 74.7 &  \colorbox{myyellow}{46.6} &  \colorbox{myyellow}{58.7} &  \colorbox{myyellow}{53.5} & 64.4 \\ \hline
\end{tabular}%
}
\end{table*}

As described in Table \ref{tab:same_question}, due to the two fundamental conditions for awakening, \(C1\) and \(C2\), the evaluation data employed varies among different LLMs (The initial dataset is the same). We conducted an intersection of questions assessed across diverse knowledge variants for the same LLM to critically analyze the influence of knowledge type on awakening performance. The original experimental results can be found in the Appendix \ref{Intersection}.

\(A_x\)-\(R_q^1\)-\(B_x\) performs steadily across the five LLMs on the 2Wiki but less effectively on CWQ, whereas \(B\)-\(R_x\)-\(A_x\) shows consistent results on both. Notably, \(A_x\)-\(R_q^1\)-\(B_x\) achieves the highest scores with Qwen2.5-32B (4.2\% Hits@1, 7.3\% Hits@10) and 6.1\% Hits@1 with Llama3.1-70B on 2Wiki. In contrast, \(B\)-\(R_x\)-\(A_x\) records 34.3\% and 46.1\% in Hits@1 and Hits@10 on CWQ using Llama3.1-70B. This success is due to the fact that \(B\)-\(R_x\)-\(A_x\) introduces more accessible second-hop information that is closer to latent knowledge \(\mathcal{K'}\), while \(A_x\)-\(R_q^1\)-\(B_x\)'s effectiveness depends on how relations are described, with 2Wiki aligning well with Wikidata and CWQ presenting more varied expressions. Furthermore, we test the performance that utilizes \(A_q\)-\(R_q^1\) to replace the \(A_q\)-\(R_q^1\)-\(B\) as the root feature in reranking stage since \(B\) is not identified for a real question, which shows the same trends when using \(A_q\)-\(R_q^1\)-\(B\). Due to page limitation, the evaluation results of knowledge type \(B_x\)-\(R_x\)-\(A_q\) and \(A_x\)-\(R_q^2\)-\(B_x\), and the performance of API-based LLMs can be found in Appendix \ref{app:res}.

\subsection{RQ2: What are the Factors that Determine the Performance of Knowledge Awakening?}

Our study primarily investigates two factors that influence knowledge awakening. More results of other factors can be found in Appendix \ref{temp}.

(1) \textbf{Model size} affects the performance of knowledge awakening. As shown in Figure \ref{different_llm}, within the Qwen2.5 series, performance improves significantly as the model scales from 7B (28 layers) to 32B (64 layers). While performance continues to improve at 72B (80 layers), the rate of improvement drops noticeably. This may be attributed to the increase in the number of layers in larger models, which can potentially hinder the effectiveness of awakening\footnote{https://qwenlm.github.io/blog/qwen2.5-llm/}.

(2) \textbf{Thinking time} can enhancing knowledge awakening. As shown in Figure \ref{fig:instruct_r1}, the DeepSeek-R1 distilled model outperforms Instruct. Qwen-32B's Instruct version with \(A_q\)-\(R_q\)-\(B\) knowledge scores 19.2\%, whereas the distilled R1 version achieves 35.4\%, showing a 16.2\% improvement. Knowledge \(\mathcal{K}\) propagates more deeply through a longer reasoning chain in the LLM, enhancing the model's ability to activate relevant internal knowledge.

\subsection{RQ3: What is the Performance of the Awakening under an Incomplete KGs?}
This section evaluates the effectiveness of awaken rule-based method in Task2. As shown in Table \ref{tab:qa}, the analyses are listed as follows:

(1) The Awaken rule-based method (ours) demonstrates substantial improvements in both Hits@1 and Hits@10 metrics across three datasets using various LLMs. For instance, on Mintaka using Llama3.1-70B-Instruct, our method boosts Hits@1 by 5.8\% and Hits@10 by 8.5\% with union knowledge (\(A_x\)-\(R_q\)-\(B_x\) and \(A_q\)-\(R_q\)-\(B_x\) (T7 and T1*)). On 2Wiki, using \(A_q\)-instance-of with Llama3.1-8B-Instruct, our method achieves state-of-the-art results, with Hits@1 and Hits@10 improving by 9.9\% and 6.4\%, respectively, showcasing its strong effectiveness under incomplete KGs. Moreover, combining all knowledge types didn't yield optimal performance, possibly due to conflicts between them.   

(2) The types of knowledge required to achieve optimal performance vary across different LLMs and datasets. Specifically, after incorporating \(A_q\)-instance-of (T4), all evaluated LLMs achieved their highest performance on the Mintaka dataset, suggesting a greater ambiguity of the entity in Mintaka questions. \(A_x\)-\(R_q\)-\(B_x\) (T7) achieve Hits@1 of 40.1\% and 54.0\% on Llama3.1-8B-Instruct and Qwen2.5-32B-Instruct, respectively, on the CWQ dataset. For Qwen2.5-32B-Instruct, combining \(A_x\)-\(R_q\)-\(B_x\) and \(A_q\)-\(R_q\)-\(B_x\) (T7 and T1*) gives the best Hits@1 scores on 2Wiki and CWQ at 41.3\% and 49.6\%. The awaken rule-based method effectively discovers external knowledge to awaken LLMs’ parametric knowledge.

(3) Question embedding methods like DiFaR and QD-DiFaR underperform compared to No RAG on Mintaka, 2Wiki and CWQ. For example, DiFaR (question) scores 25.9\% Hits@1, which is 25.4\% lower than No RAG on Mintaka with Llama3.1-8B-Instruct, highlighting the limitations of question similarity-based retrieval when KGs are incomplete. Additionally, DiFaR (entity) and QD-DiFaR (entity) achieve 23.9\% and 27.9\% Hits@1 with Qwen-32B, below the question embedding methods' performance of 32.5\% and 35.8\%. This is due to entities in questions often failing to match in KGs, creating a knowledge gap between queries and KGs.

\section{Conclusion}

This work is rooted in the theory of spreading activation in the human brain and provides both 
theoretical and empirical analysis of awakening LLMs by reinjecting partially relevant knowledge already embedded in their parameters. 
We demonstrate the existence of this phenomenon through a series of experiments, revealing how different types of knowledge and external factors influence the awakening effect. Furthermore, we simulate a realistic setting with incomplete knowledge bases and introduce the Unseen Entity QA task. Experimental results show that retrieval guided by awakening-based rules outperforms existing methods, highlighting the practical value of knowledge awakening in real-world scenarios. 

Despite the improved effectiveness of RAG under incomplete knowledge settings through our awaken rule-based approach, this work still has several limitations. First, due to the complexity of relationships and entities in documents, rule-based methods are currently not applicable to document-level knowledge retrieval. Second, the knowledge extracted via predefined rules may be incomplete; exploring more automated and in-depth ways to mine richer knowledge remains an important future direction. Finally, the capacity of LLMs to awaken internal knowledge requires further investigation, which could offer new insights into the interpretability of LLMs.

\bibliographystyle{unsrt}
\bibliography{awaken}

\appendix
\newpage
\twocolumn

\section{The Theoretical Proof That LLMs Are Awakenable}
\label{app:proof}
According to the equation \ref{eq2}, the attention mechanism is:
\begin{equation}
    A(q_i,K,V)=\sum_{j=1}^n a_{ij}v_j,
\end{equation}
where
\begin{equation}
    \alpha_{ij}=\frac{\mathrm{exp}(q_i \cdot k_j/\sqrt{d})}{\sum_{j=1}^n\mathrm{exp}(q_i \cdot k_j/\sqrt{d})}.
\end{equation}
According to it and the equation \ref{eq5}, we have:
\begin{equation}
     \mathbf{H}^{(L)}= \textstyle \prod_{l=1}^{L}  \mathbf{a}_{ij}^{(l)}.
     \label{9}
\end{equation}
Due to \(\mathbf{H}^{(L)}\) being computable, we define the path \(\pi=(i_0,i_1,...,i_L)\) represents the input token \(x_{i_0}=(\mathcal{Q,K})\), through the multi-layer Transformer activates \(H_{i_L}\)\(\mathcal{K'}\). In light of this, the path probability functional can be defined as:
\begin{equation}
    \mathcal{F}[\pi]:=
    \mathrm{log}\mathbf{H}^{(L)}=\sum_{l=1}^{L}\mathrm{log}\mathbf{a}_{i_{l-1},i_l}^{(l)}.
\end{equation}
Combine the equation \ref{9}, we have:
\begin{equation}
    \mathcal{F}[\pi]:=\mathrm{log}\mathbf{a}_{i_{l-1},i_l}^{(l)}=\frac{\left \langle  \mathbf{q}_{i_{l-1}}^{(l)},\mathbf{k}_{i_l}^{(l)}\right \rangle}{\sqrt d}-\mathrm{log}\sum_{j=1}^{n}\mathrm{exp}(\frac{\left \langle \mathbf{q}_{i_{l-1}}^{(l)},\mathbf{k}_{j}^{(l)}\right \rangle}{\sqrt d}),
\end{equation}
it can be simplified into:
\begin{equation}
    \mathcal{F}[\pi]:=\mathrm{log}\mathbf{a}_{i_{l-1},i_l}^{(l)}=\mathbf{a}_{i_{l-1},i_l}^{(l)}-\mathrm{log}\sum_{j=1}^{n}\mathrm{exp}(\mathbf{a}_{i_{l-1},j}^{(l)}),
\end{equation}
which is a standard Log-Softmax function. The Hessian of this Log-Softmax function is
\begin{align}
    & \frac{\partial^2}{\partial \mathbf{a}_{i_{(l-1)},i_l} \partial \mathbf{a}_{i_{(l-1)},j}} \text{log-softmax}(\mathbf a_{i_{(l-1)},i_l}) \notag \\ =
& \begin{cases}
\mathbf{a}_{i_{(l-1)},i_l} - \mathbf{a}_{i_{(l-1)},i_l}^2 & \text{if } j = k \\
\mathbf{a}_{i_{(l-1)},i_l} \cdot \mathbf{a}_{i_{(l-1)},j} & \text{if } j \ne k.
\end{cases}
\end{align}
This matrix is negative semi-definite, implying that the Log-Softmax function \(\mathcal{F}\) is concave. Since the domain of this functional is a combinatorial space of finite length, the extreme value theorem ensures that \(\mathcal{F}\) attains a global maximum within its domain. \textbf{This further implies that there exists a point along the path at which \(\mathcal{F}\) reaches its maximal probability}. Consequently, there may exist an external knowledge state \(\mathcal{K}\) that increases the probability of \(\mathbb{P}\) thereby facilitating the awakening of LLMs.

\section{The Others Experimental Results}
\label{app:res}

\subsection{Original Awakening Results without Intersection}
\label{Intersection}

\begin{table*}[t]
\small
\centering
\caption{The awaken experimental results on two datasets across 5 LLMs. ``T'' represents the knowledge types (e.g., No RAG acts the knowledge \( \ X - R - Y  \)), ``F'' indicates the number of samples be used to test. \colorbox{red!10}{\textcolor{red!10}{\rule{1em}{0.8ex}}} represents No RAG results, \colorbox{green!10}{\textcolor{green!10}{\rule{1em}{0.7ex}}} indicates the awaken results.}
\label{tab:original}
\setlength{\tabcolsep}{0.055cm}{%
\begin{tabular}{cccccccccccccccc}
\toprule
 \multicolumn{1}{c}{\multirow{3}{*}{\textbf{Knowledge type}}} & \multicolumn{3}{c}{\textbf{Llama3.1-8B}} & \multicolumn{3}{c}{\textbf{Llama3.1-70B}} & \multicolumn{3}{c}{\textbf{Qwen2.5-7B}} & \multicolumn{3}{c}{\textbf{Qwen2.5-32B}} & \multicolumn{3}{c}{\textbf{Qwen2.5-72B}}  \rule{0pt}{2ex}\\ \cline{2-16} 
 & F & Hits@1 & Hits@10 & F & Hits@1 & Hits@10 & F & Hits@1 & Hits@10 & F & Hits@1 & Hits@10 & F & Hits@1 & Hits@10  \rule{0pt}{2ex}\\ \cline{2-16} 
 \multicolumn{1}{c}{} & \multicolumn{15}{c}{\textbf{Base}}  \rule{0pt}{2ex}\\ \cline{2-16} 
 & \multicolumn{15}{c}{\textbf{2Wiki}}  \rule{0pt}{2ex}\\ 

\rowcolor{red!10}
No RAG & 355 & 0.0 & 14.0  & 892 & 0.0 & 16.6 & 448 & 0.0 & 12.5  & 510 & 0.0 & 12.7  & 568 & 0.0 & 22.2   \rule{0pt}{2ex}\\
\rowcolor{green!10}
\textbf{\textbf{\(A_q\)-\(R_q^1\)-\(B\)}} & 355 & 2.3  & 11.0  & 892 & 0.6  & 5.2  & 448 & 4.7  & 18.8  & 510 & 5.1  & 20.0  & 568 & 7.7  & 29.0 \\
\rowcolor{green!10}
\textbf{\textbf{\(A_q\)-\(R_x\)-\(B_x\)}} & 197 & 2.0  & 7.6  & 663 & 0.2  & 5.3  & 286 & 1.4  & 11.9  & 374 & 2.7  & 12.3  & 432 & 2.8  & 22.7  \\
\rowcolor{green!10}
\textbf{\textbf{\(B\)-\(R_x\)-\(A_x\)}} & 243 & 3.3 &10.7  & 799 & 0.6  & 5.6  & 361 & 3.9  & 19.1  & 416 & 6.7  & 21.6 & 473 & 7.2  & 29.4  \\
\rowcolor{green!10}
\textbf{\textbf{\(A_x\)-\(R_q^1\)-\(B_x\)}} & 254 &2.8  & 13.4  & 683 & 1.0 & 5.4  & 321 & 5.3  & 16.8  & 409 & 5.9  & 18.6  & 453 & 6.0  & 26.3   \\
\rowcolor{green!10}
\textbf{\textbf{\(A_q\)-\(R_x\)-\(Z\)-\(R_x\)-\(B\)}} & 128 & 3.9  & 11.7  & 448 & 0.7  & 6.2  & 164 & 5.5  & 20.1  & 193 & 6.7  & 24.4  & 271 & 5.9  & 25.8  \\
\rowcolor{green!10}
\(A_q\)-\(\mathrm{instance}\)-\(\text{of}\) & 188 & 3.2  & 11.2 & 609 & 0.3 & 3.8  & 266 & 2.6  & 16.2  & 375 & 3.5  & 11.7  & 414 & 3.9  & 21.0  \\ \cline{2-16} 
 & \multicolumn{15}{c}{\textbf{CWQ}}  \rule{0pt}{2ex}\\
 \rowcolor{red!10}
No RAG & 186 & 0.0 & 40.9 & 185 & 0.0 & 45.9  & 148 & 0.0 & 41.9  & 163 & 0.0 & 46.0  & 163 & 0.0 & 52.1  \rule{0pt}{2ex}\\
\rowcolor{green!10}
\textbf{\textbf{\(A_q\)-\(R_q^1\)-\(B\)}} & 186 & 36.6 & 73.1 & 185 & 9.2  & 48.1  & 148 & 24.3  & 60.1  & 163 & 35 & 68.1 & 163 & 30.1  & 64.4  \\
\rowcolor{green!10}
\textbf{\textbf{\(A_q\)-\(R_x\)-\(B_x\)}} & 166 & 30.1 & 66.9 & 177 &6.2  & 44.1  & 136 & 29.4  & 62.5  & 157 & 33.1  & 66.9  & 158 & 31.6  & 64.6  \\
\rowcolor{green!10}
\textbf{\textbf{\(B\)-\(R_x\)-\(A_x\)}} &  161 &  37.9 &  70.8 &  174 &  5.7  & 40.8  &  139 & 25.9  &  61.9 &  149 & 36.2  & 69.1  &  155 &  34.2  &  61.3  \\
\rowcolor{green!10}
\textbf{\textbf{\(A_x\)-\(R_q^1\)-\(B_x\)}} &  171 & 24.6 & 56.1 & 167 &  4.2  &  26.3  & 132 & 15.2  & 40.9  &  150 & 22.0  & 46.0  &  155 & 18.7 & 53.5 \\
\rowcolor{green!10}
\textbf{\textbf{\(A_q\)-\(R_x\)-\(Z\)-\(R_x\)-\(B\)}} & 129 &  41.9 & 71.3 & 114 & 9.6  & 38.6  & 84 & 21.4  & 52.4  & 86 & 33.7  & 64.0  & 115 & 30.4  &  58.3 \\
\rowcolor{green!10}
\(A_q\)-\(\mathrm{instance}\)-\(\text{of}\) & 111 & 33.3 & 70.3 & 160 & 5.6  & 38.8  & 125 & 25.6  & 68.0  & 135 & 29.6  & 64.4  & 136 & 33.8  & 66.2 \\ \cline{2-16} 
 & \multicolumn{15}{c}{\textbf{Instruct}}  \rule{0pt}{2ex}\\ \cline{2-16} 
 & \multicolumn{15}{c}{\textbf{2Wiki}}  \rule{0pt}{2ex}\\
 \rowcolor{red!10}
No RAG & 1668 & 0.0  & 10.6  & 1443 &0.0 & 15.7  & 946 & 0.0 & 3.9 & 934 & 0.0 & 3.5 & 1180 & 0.0 & 3.8  \rule{0pt}{2ex}\\
\rowcolor{green!10}
\textbf{\textbf{\(A_q\)-\(R_q^1\)-\(B\)}} & 1668 & 4.1  & 14.2  & 1443 &5.9 &5.9& 964 & 3.7 & 7.2 & 934 & 6.3 & 12.6 & 1180  & 7.0 & 11.9 \\
\rowcolor{green!10}
\textbf{\textbf{\(A_q\)-\(R_x\)-\(B_x\)}} & 1358 &4.2 & 4.2  &1194 & 4.5  & 14.1  & 760 &2.4&4.6& 735 &3.5&7.6& 945 &4.9&  9.4\\
\rowcolor{green!10}
\textbf{\textbf{\(B\)-\(R_x\)-\(A_x\)}} & 1643 &3.6 & 16.0  &1421 & 5.6  & 17.6  &935&3.1&8.3&882&5.7&12.8&1148& 7.3 &  13.6\\
\rowcolor{green!10}
\textbf{\textbf{\(A_x\)-\(R_q^1\)-\(B_x\)}} & 1282 & 4.1  & 13.2  & 1118  & 4.7  & 13.1  &718&3.9&6.4&677&5.3&10.5&931&6.1&  11.1\\
\rowcolor{green!10}
\textbf{\textbf{\(A_q\)-\(R_x\)-\(Z\)-\(R_x\)-\(B\)}} &890  & 4.5  &13.0 & 133.0  &5.3 & 12.0  &362&1.7&5.0&422&6.4&11.1&357&8.4&12.9\\
\rowcolor{green!10}
\(A_q\)-\(\mathrm{instance}\)-\(\text{of}\) & 1357 & 4.0  & 10.6  & 1172  & 4.2  & 19.0  &636&2.8&6.4&599&2.0&6.3&895&4.4&9.4\\ \cline{2-16} 
 & \multicolumn{15}{c}{\textbf{CWQ}}  \rule{0pt}{2ex}\\
 \rowcolor{red!10}
No RAG & 467 & 0.0 & 30.2 & 390 & 0.0 & 19.0 & 387 & 0.0 & 12.1 & 316 & 0.0 & 9.5 & 313 & 0.0 & 8.6  \rule{0pt}{2ex}\\
\rowcolor{green!10}
\textbf{\textbf{\(A_q\)-\(R_q^1\)-\(B\)}} & 467 & 39.4 & 56.7 & 390 & 33.3 & 44.6 & 387 & 25.3 & 38.2 & 316 & 37.3 & 46.2 & 313 & 28.1 & 35.1 \\
\rowcolor{green!10}
\textbf{\textbf{\(A_q\)-\(R_x\)-\(B_x\)}} & 467 & 33.8 & 54.8 & 390 & 32.6 & 45.1 & 387 & 26.1 & 38.5 & 312 & 35.3 & 46.2 & 311 & 29.6 & 35.7 \\
\rowcolor{green!10}
\textbf{\textbf{\(B\)-\(R_x\)-\(A_x\)}} & 467 & 37.0 & 58.2 & 389 & 34.2 & 46.0 & 384 & 27.6 & 38.3 & 314 & 37.6 & 49.4 & 309 & 31.1 & 36.9 \\
\rowcolor{green!10}
\textbf{\textbf{\(A_x\)-\(R_q^1\)-\(B_x\)}} & 467 & 16.3 & 38.8 & 390 & 12.1 & 24.1 & 387 & 7.0 & 15.2 & 313 & 16.3 & 31.6 & 311 & 16.4 & 27.3 \\
\rowcolor{green!10}
\textbf{\(A_q\)-\(R_x\)-\(Z\)-\(R_x\)-\(B\)} & 446 & 34.3 & 54.3 &376 & 33.8 &  46.0 & 357 & 20.4 & 30.8 &  262 & 32.1 & 43.9 & 273 & 30.0 & 33.7 \\
\rowcolor{green!10}
\(A_q\)-\(\mathrm{instance}\)-\(\text{of}\) & 467 & 33.6 & 58.2 & 389 & 32.9 & 45.2 & 385 & 27.8 & 40.3 & 315 & 35.6 & 49.2 & 312 & 26.6 & 35.3 \\ \bottomrule
\end{tabular}%
}
\end{table*}

As shown in Table \ref{tab:original}, the LLMs were successfully awakened, with Instruct-tuned models demonstrating more stable behavior. In contrast, Base models—lacking the Instruct-tuning process—exhibited less controllable outputs during knowledge probing and question answering, often producing undesired responses, particularly on Llama3.1-70B. Since LLMs possess varying amounts of different types of knowledge during probing, we adopt an intersection-based evaluation strategy to enable a fair comparison of awakening effects across knowledge types. Specifically, we evaluate only on the shared subset of samples that were successfully awakened across different probing settings. The complete results are presented in Table \ref{tab:same_question_appendix}.

Specifically, on Llama3.1-8B-Base, awakening knowledge \(A_q\)-\(R_q^1\)-\(B\) achieves 2.3\% Hits@1 and 11.0\% Hits@10, outperforming No RAG in Hits@1 but underperforming it in Hits@10 (3.0\%) on 2Wiki. This pattern is also clearly observed on Llama3.1-70B-Base, yet it is nearly absent in the Qwen2.5-Base series. This discrepancy may be attributed to differences in the pretraining data for Qwen. This phenomenon is notably mitigated in the Instruct versions of LLMs. Instruct-tuned models demonstrate improved capabilities in leveraging internal knowledge and following user instructions, leading to more pronounced awakening effects and a greater number of valid filtered samples. For instance, Llama3.1-70B-Instruct achieves 5.6\% Hits@1 and 17.6\% Hits@10 on 1,421 samples when injected with \(B\)-\(R_x\)-\(A_x\). These results indicate that there remains substantial room for improvement in how models utilize their internal knowledge.

\subsection{The Awaken Performance of Base-version LLM}

\begin{table*}[t]
\small
\centering
\caption{The awaken experimental results on two datasets across 5 LLMs. ``F'' indicates the number of samples be used to test, which after filter in section \ref{4.2}. \colorbox{red!10}{\textcolor{red!10}{\rule{1em}{0.8ex}}} represents No RAG results, \colorbox{green!10}{\textcolor{green!10}{\rule{1em}{0.7ex}}} indicates the awaken results. The best two results are highlighted by \colorbox{orange!60}{1st} and \colorbox{myyellow}{2nd}. Fewer samples remain after two-stage filtering in base models due to the lack of ability of instruction following.}
\label{tab:same_question_appendix}
\setlength{\tabcolsep}{0.06cm}{%
\begin{tabular}{ccclcclcclcclccl}
\toprule
\multicolumn{1}{c}{\multirow{3}{*}{\textbf{Knowledge type}}} & \multicolumn{3}{c}{\textbf{Llama3.1-8B}} & \multicolumn{3}{c}{\textbf{Llama3.1-70B}} & \multicolumn{3}{c}{\textbf{Qwen2.5-7B}} & \multicolumn{3}{c}{\textbf{Qwen2.5-32B}} & \multicolumn{3}{c}{\textbf{Qwen2.5-72B}} \rule{0pt}{2ex}\\ \cline{2-16}
\multicolumn{1}{c}{} & F & Hits@1 & Hits@10 & F & Hits@1 & Hits@10 & F & Hits@1 & Hits@10 & F & Hits@1 & Hits@10 & F & Hits@1 & Hits@10 \rule{0pt}{2ex}\\ \cline{2-16} 
\multicolumn{1}{c}{} & \multicolumn{15}{c}{\textbf{Base}} \rule{0pt}{2ex}\\ \cline{2-16} 
 & \multicolumn{15}{c}{\textbf{2Wiki}} \rule{0pt}{2ex}\\ \cline{2-16}
 \rowcolor{red!10}
No RAG & 49 & 0.0 & \cellcolor{myyellow}{\centering 10.2} & 351 & 0.0 & \cellcolor{orange!60}{\centering 21.4} & 101 & 0.0 & 13.9 & 150 & 0.0 & 14.7 & 214 & 0.0 & 28.5 \rule{0pt}{2ex}\\
\rowcolor{green!10}
\textbf{\(A_q\)-\(R_x\)-\(B\)} & 49 & \cellcolor{orange!60}{\centering 2.0} & \cellcolor{orange!60}{\centering 12.2} & 351 & 0.3 & 5.7 & 101 & \cellcolor{myyellow}{\centering 4.0} & 17.8 & 150 & \cellcolor{myyellow}{\centering 5.3} & 20.7 & 214 & 7.0 & 29.0 \\
\rowcolor{green!10}
\textbf{\(A_q\)-\(R_x\)-\(B_x\)} & 49 & \cellcolor{orange!60}{\centering 2.0} & 8.2 & 351 & 0.3 & 6.0 & 101 & 1.0 & 10.9 & 150 & 3.3 & 16.0 & 214 & 3.3 & 22.0 \\
\rowcolor{green!10}
\textbf{\(B\)-\(R_x\)-\(A_x\)} & 49 & \cellcolor{orange!60}{\centering 2.0} & \cellcolor{myyellow}{\centering  10.2} & 351 & \cellcolor{myyellow}{\centering 0.9} & \cellcolor{myyellow}{\centering 6.6} & 101 & \cellcolor{myyellow}{\centering 4.0} & \cellcolor{orange!60}{\centering 20.8} & 150 & 6.0 & \cellcolor{myyellow}{\centering 22.7} & 214 & \cellcolor{orange!60}{\centering 9.8} & \cellcolor{myyellow}{\centering 32.2} \\
\rowcolor{green!10}
\textbf{\(A_x\)-\(R_q^1\)-\(B_x\)} & 49 & 0.0 & \cellcolor{myyellow}{\centering  10.2} & 351 & \cellcolor{orange!60}{\centering 1.4} & 5.4 & 101 & \cellcolor{orange!60}{\centering 5.0} & 16.8 & 150 & \cellcolor{orange!60}{\centering 7.3} & \cellcolor{orange!60}{\centering 24.0} & 214 & \cellcolor{myyellow}{\centering 7.5} & \cellcolor{orange!60}{\centering 32.7} \\
\rowcolor{green!10}
\textbf{\(A_q\)-\(R_x\)-\(Z\)-\(R_x\)-\(B\)} & 49 & \cellcolor{orange!60}{\centering 2.0} & 6.1 & 351 & \cellcolor{myyellow}{\centering 0.9} & 4.8 & 101 & \cellcolor{myyellow}{\centering 4.0} & \cellcolor{myyellow}{\centering 18.8} & 150 & \cellcolor{orange!60}{\centering 7.3} & \cellcolor{orange!60}{\centering 24.0} & 214 & 7.0 & 26.2 \\
\rowcolor{green!10}
\(A_q\)-\(\mathrm{instance}\)-\(\text{of}\) & 49 & \cellcolor{orange!60} 2.0 & 6.1 & 351 & 0.3 & 3.1 & 101 & 2.0 & 15.8 & 150 & \cellcolor{myyellow}{\centering 5.3} & 15.3 & 214 & 3.7 & 21.5 \\ \cline{2-16} 
 & \multicolumn{15}{c}{\textbf{CWQ}} \rule{0pt}{2ex}\\ \cline{2-16}
 \rowcolor{red!10}
No RAG & 54 & 0.0 & 50.0 & 83 & 0.0 & \cellcolor{orange!60}{\centering 44.6} & 59 & 0.0 & 40.7 & 66 & 0.0 & 48.5 & 89 & 0.0 & 53.9 \rule{0pt}{2ex}\\
\rowcolor{green!10}
\textbf{\(A_q\)-\(R_x\)-\(B\)} & 54 & \cellcolor{orange!60}{\centering 40.7} & \cellcolor{myyellow}{\centering 70.4} & 83 & \cellcolor{orange!60}{\centering 12.0} & \cellcolor{orange!60}{\centering 44.6} & 59 & \cellcolor{myyellow}{\centering 25.4} & 45.8 & 66 & 31.8 & 60.6 & 89 & 30.3 & \cellcolor{myyellow}{\centering 67.4} \\
\rowcolor{green!10}
\textbf{\(A_q\)-\(R_x\)-\(B_x\)} & 54 & 25.9 & 64.8 & 83 & 3.6 & 37.3 & 59 & \cellcolor{orange!60}{\centering 28.8} & \cellcolor{myyellow}{\centering 54.2} & 66 & 28.8 & 59.1 & 89 & \cellcolor{myyellow}{\centering 33.7} & \cellcolor{myyellow}{\centering 67.4} \\
\rowcolor{green!10}
\textbf{\(B\)-\(R_x\)-\(A_x\)} & 54 & \cellcolor{myyellow}{\centering 33.3} & 66.7 & 83 & 3.6 & 36.1 & 59 & 22.0 & 52.5 & 66 & \cellcolor{orange!60}{\centering 34.8} & \cellcolor{orange!60}{\centering 66.7} & 89 & 28.1 & 61.8 \\
\rowcolor{green!10}
\textbf{\(A_x\)-\(R_q^1\)-\(B_x\)} & 54 & \cellcolor{myyellow}{\centering 33.3} & 63.0 & 83 & 4.8 & 24.1 & 59 & 10.2 & 35.6 & 66 & 25.8 & 51.5 & 89 & 20.2 & 56.2 \\
\rowcolor{green!10}
\textbf{\(A_q\)-\(R_x\)-\(Z\)-\(R_x\)-\(B\)} & 54 & \cellcolor{orange!60}{\centering 40.7} & \cellcolor{orange!60}{\centering 74.1} & 83 & \cellcolor{orange!60}{\centering 12.0} & \cellcolor{myyellow}{\centering 39.8} & 59 & 15.3 & 50.8 & 66 & 31.8 & 60.6 & 89 & 30.3 & 58.4 \\
\rowcolor{green!10}
\(A_q\)-\(\mathrm{instance}\)-\(\text{of}\) & 54 & \cellcolor{myyellow}{\centering 33.3} & 68.5 & 83 & \cellcolor{myyellow}{\centering 9.6} & 37.3 & 59 & 18.6 & \cellcolor{orange!60}{\centering 57.6} & 66 & \cellcolor{myyellow}{\centering 33.3} & \cellcolor{myyellow}{\centering 63.6} & 89 & \cellcolor{orange!60}{\centering 36.0} & \cellcolor{orange!60}{\centering 69.7} \\ \cline{2-16} 
 & \multicolumn{15}{c}{\textbf{Instruct}} \rule{0pt}{2ex}\\ \cline{2-16} 
 & \multicolumn{15}{c}{\textbf{2Wiki}} \rule{0pt}{2ex}\\ \cline{2-16}
 \rowcolor{red!10}
No RAG & 815 & 0.0 & 13.3 & 693 & 0.0 & 16.2 & 289 & 0.0 & 4.8 & 323 & 0.0 & 4.3 & 306 & 0.0 & 3.9 \rule{0pt}{2ex}\\
\rowcolor{green!10}
\textbf{\(A_q\)-\(R_x\)-\(B\)} & 815 & 3.7 & \cellcolor{myyellow}{\centering 15.5} & 693 & \cellcolor{myyellow}{\centering 5.9} & \cellcolor{myyellow}{\centering 17.6} & 289 & \cellcolor{myyellow}{\centering 2.8} & 5.2 & 323 & 5.3 & 13.0 & 306 & 4.9 & 11.1 \\
\rowcolor{green!10}
\textbf{\(A_q\)-\(R_x\)-\(B_x\)} & 815 & \cellcolor{myyellow}{\centering 4.4} & 14.8 & 693 & 4.6 & 15.6 & 289 & 2.4 & 4.5 & 323 & 4.3 & 10.2 & 306 & 6.9 & 11.1 \\
\rowcolor{green!10}
\textbf{\(B\)-\(R_x\)-\(A_x\)} & 815 & 3.1 & \cellcolor{orange!60}{\centering 16.4} & 693 & 5.8 & 17.5 & 289 & 1.4 & \cellcolor{myyellow}{\centering 5.5} & 323 & 5.6 & \cellcolor{myyellow}{\centering 13.6} & 306 & 6.2 & \cellcolor{orange!60}{\centering 13.7} \\
\rowcolor{green!10}
\textbf{\(A_x\)-\(R_q^1\)-\(B_x\)} & 815 & 3.3 & 13.5 & 693 & \cellcolor{orange!60}{\centering 6.1} & 15.7 & 289 & \cellcolor{orange!60}{\centering 4.2} & \cellcolor{orange!60}{\centering 7.3} & 323 & \cellcolor{myyellow}{\centering 6.2} & \cellcolor{orange!60}{\centering 13.9} & 306 & \cellcolor{myyellow}{\centering 6.5} & \cellcolor{myyellow}{\centering 11.8} \\
\rowcolor{green!10}
\textbf{\(A_q\)-\(R_x\)-\(Z\)-\(R_x\)-\(B\)} & 815 & 4.0 & 12.6 & 693 & 5.8 & 15.9 & 289 & 1.7 & 4.5 & 323 & \cellcolor{orange!60}{\centering 6.8} & 11.1 & 306 & \cellcolor{orange!60}{\centering 8.8} & \cellcolor{orange!60}{\centering 13.7} \\
\rowcolor{green!10}
\(A_q\)-\(\mathrm{instance}\)-\(\text{of}\) & 815 & \cellcolor{orange!60}{\centering 4.7} & 12.9 & 693 & 4.6 & \cellcolor{orange!60}{\centering 19.8} & 289 & 1.0 & 3.1 & 323 & 3.4 & 8.4 & 306 & 4.6 & 10.5 \\ \cline{2-16} 
 & \multicolumn{15}{c}{\textbf{CWQ}} \rule{0pt}{2ex}\\ \cline{2-16} 
 \rowcolor{red!10}
No RAG & 446 & 0.0 & 30.7 & 375 & 0.0 & 19.5 & 353 & 0.0 & 12.5 & 256 & 0.0 & 9.0 & 270 & 0.0 & 8.9\rule{0pt}{2ex} \\
\rowcolor{green!10}
\textbf{\(A_q\)-\(R_x\)-\(B\)} & 446 & \cellcolor{orange!60}{\centering 39.2} & 56.3 & 375 & 33.3 & 45.1 & 353 & 25.2 & \cellcolor{myyellow}{\centering 38.5} & 256 & \cellcolor{myyellow}{\centering 37.1} & 45.7 & 270 & 25.9 & 33.3 \\
\rowcolor{green!10}
\textbf{\(A_q\)-\(R_x\)-\(B_x\)} & 446 & 34.1 & 54.9 & 375 & 32.8 & 45.3 & 353 & 26.1 & 38.2 & 256 & \cellcolor{myyellow}{\centering 37.1} & 46.1 & 270 & 28.9 & \cellcolor{myyellow}{\centering 35.2} \\
\rowcolor{green!10}
\textbf{\(B\)-\(R_x\)-\(A_x\)} & 446 & \cellcolor{myyellow}{\centering 37.0} & \cellcolor{orange!60}{\centering 58.7} & 375 & \cellcolor{orange!60}{\centering 34.4} & \cellcolor{orange!60}{\centering 46.1} & 353 & \cellcolor{myyellow}{\centering 27.5} & 37.7 & 256 & \cellcolor{orange!60}{\centering 38.7} & \cellcolor{orange!60}{\centering 50.4} & 270 & \cellcolor{myyellow}{\centering 29.6} & \cellcolor{orange!60}{\centering 35.6} \\
\rowcolor{green!10}
\textbf{\(A_x\)-\(R_q^1\)-\(B_x\)} & 446 & 16.4 & 39.0 & 375 & 12.5 & 24.3 & 353 & 7.6 & 16.4 & 256 & 16.4 & 32.4 & 270 & 17.8 & 28.5 \\
\rowcolor{green!10}
\textbf{\(A_q\)-\(R_x\)-\(Z\)-\(R_x\)-\(B\)} & 446 & 34.3 & 54.3 & 375 & \cellcolor{myyellow}{\centering 33.9} & \cellcolor{myyellow}{\centering 45.9} & 353 & 20.4 & 30.9 & 256 & 32.4 & 44.5 & 270 & \cellcolor{orange!60}{\centering 30.4} & 34.1 \\
\rowcolor{green!10}
\(A_q\)-\(\mathrm{instance}\)-\(\text{of}\) & 446 & 33.4 & \cellcolor{myyellow}{\centering 58.3} & 375 & 33.1 & 45.3 & 353 & \cellcolor{orange!60}{\centering 28.6} & \cellcolor{orange!60}{\centering 41.1} & 256 & 35.5 & \cellcolor{myyellow}{\centering 48.8} & 270 & 24.8 & 34.1 \\ \bottomrule
\end{tabular}%
}
\end{table*}

As shown in Table \ref{tab:same_question_appendix}, we further intersect the awakening results of Base version LLMs to analyze the influence of different knowledge on model performance. Overall, the Llama3.1 series performs poorly on 2Wiki, while Qwen models remain consistently stable. On the CWQ dataset, however, all models exhibit strong performance. For example, after injecting \(B\)-\(R_x\)-\(A_x\), Qwen2.5-32B-Base achieves 34.8\% Hits@1 and 66.7\% Hits@10. Moreover, the overall trend differs slightly from that observed in the Instruct versions: Base models tend to prefer knowledge located on the gold reasoning path, while Instruct models exhibit better reasoning ability when dealing with non-gold knowledge. This phenomenon may stem from co-occurrence patterns of knowledge in the training data.

\subsection{The Awakening Results on API-based LLMs}
\label{api}
\begin{table}[t]
\centering
\caption{Performance on API-based LLMs. We evaluate two models: DeepSeek-V3-250324 and GPT4.1-mini. The DeepSeek API limits candidate returns to five, so we use the Hits@5 metric.}
\label{tab:api}
\setlength{\tabcolsep}{0.083cm}{
\begin{tabular}{ccccccc}
\hline
\multirow{3}{*}{\begin{tabular}[c]{@{}c@{}}Knowledge\\ Type\end{tabular}} & \multicolumn{6}{c}{\textbf{DeepSeek-V3}} \\ \cline{2-7} 
 & F & Hits@1 & Hits@5 & F & Hits@1 & Hits@5 \\ \cline{2-7} 
 & \multicolumn{3}{c}{\textbf{2Wiki}} & \multicolumn{3}{c}{\textbf{CWQ}} \\ \cline{2-7} 
No RAG & 861 & 0.0 & 4.6 & 336 & 0.0 & 10.1 \\
\(A_q\)-\(R_q^1\)-\(B\) & 861 & 14.1 & 23.1 & 336 & 28.9 & 31.5 \\
\(A_q\)-\(R_x\)-\(B_x\) & 861 & 13.0 & 22.1 & 336 & 29.5 & 34.5 \\
\(B\)-\(R_x\)-\(A_x\) & 861 & 16.1 & 25.9 & 336 & 31.5 & 34.5 \\
\(A_x\)-\(R_q^1\)-\(B_x\) & 861 & 15.4 & 23.3 & 336 & 16.1 & 22.3 \\
\(A_q\)-\(R_x\)-\(Z\)-\(R_x\)-\(B\) & 861 & 16.0 & 24.7 & 336 & 30.4 & 35.1 \\
\(A_q\)-instance-of & 861 & 10.7 & 20.8 & 336 & 28.3 & 32.1 \\ \cline{2-7} 
\multicolumn{1}{l}{} & \multicolumn{6}{c}{\textbf{GPT4.1-mini}} \\ \cline{2-7} 
\multicolumn{1}{l}{} & F & Hits@1 & Hits@10 & F & Hits@1 & Hits@10 \\
No RAG & 708 & 0.0 & 7.2 & 302 & 0.0 & 14.2 \\
\(A_q\)-\(R_q\)-\(B\) & 708 & 12.1 & 23.3 & 302 & 37.7 & 44.0 \\
\(A_q\)-\(R_x\)-\(B_x\) & 708 & 9.5 & 19.4 & 302 & 36.1 & 40.1 \\
\(B\)-\(R_x\)-\(A_x\) & 708 & 8.9 & 20.1 & 302 & 37.4 & 43.7 \\
\(A_x\)-\(R_q^1\)-\(B_x\) & 708 & 9.2 & 16.9 & 302 & 18.5 & 24.8 \\
\(A_q\)-\(R_x\)-\(Z\)-\(R_x\)-\(B\) & 708 & 9.0 & 19.5 & 302 & 38.1 & 45.0 \\
\(A_q\)-instance-of & 708 & 8.6 & 26.6 & 302 & 38.1 & 43.0 \\ \hline
\end{tabular}%
}
\end{table}

\begin{table}[t]
\centering
\caption{Awakening results on Llama3.1-8B-Instruct after conducting knowledge probing via QA and NLI tasks, respectively.}
\label{tab:qa_nli}
\setlength{\tabcolsep}{0.053cm}{
\begin{tabular}{ccccccc}
\hline
\multirow{2}{*}{\begin{tabular}[c]{@{}c@{}}Knowledge\\ Type\end{tabular}} & \multicolumn{3}{c}{QA-probing} & \multicolumn{3}{c}{NLI-probing} \\ \cline{2-7} 
 & F & Hits@1 & Hits@10 & F & Hits@1 & Hits@10 \\ \cline{2-7} 
\multicolumn{1}{l}{} & \multicolumn{6}{c}{Original} \\ \cline{2-7} 
No RAG & 110 & 0.0 & 20.0 & 446 & 0.0 & 30.7 \\
\(A_q\)-\(R_q\)-\(B\) & 110 & 33.6 & 48.2 & 446 & 39.2 & 56.3 \\
\(A_q\)-\(R_x\)-\(B_x\) & 67 & 23.9 & 41.8 & 446 & 34.1 & 54.9 \\
\(B\)-\(R_x\)-\(A_x\) & 22 & 45.5 & 59.1 & 446 & 37.0 & 58.7 \\
\(A_x\)-\(R_q^1\)-\(B_x\) & 49 & 20.4 & 34.7 & 446 & 16.4 & 39.0 \\
\(A_q\)-\(R_x\)-\(Z\)-\(R_x\)-\(B\) & 27 & 29.6 & 44.4 & 446 & 34.3 & 54.3 \\
\(A_q\)-instance-of & 16 & 12.5 & 31.2 & 446 & 33.4 & 58.3 \\ \cline{2-7} 
\multicolumn{1}{l}{} & \multicolumn{6}{c}{Intersection} \\ \cline{2-7} 
No RAG & 81 & 0 & 14.8 & 81 & 0 & 14.8 \\
\(A_q\)-\(R_q\)-\(B\) & 81 & 29.6 & 43.2 & 81 & 30.9 & 44.4 \\
\(A_q\)-\(R_x\)-\(B_x\) & 49 & 18.4 & 38.8 & 49 & 20.5 & 40.8 \\
\(B\)-\(R_x\)-\(A_x\) & 14 & 35.7 & 42.9 & 14 & 35.7 & 42.9 \\
\(A_x\)-\(R_q^1\)-\(B_x\) & 38 & 21.1 & 28.9 & 38 & 13.2 & 42.1 \\
\(A_q\)-\(R_x\)-\(Z\)-\(R_x\)-\(B\) & 20 & 20 & 30 & 20 & 25 & 35 \\
\(A_q\)-instance-of & 9 & 22.2 & 44.4 & 9 & 22.2 & 44.4 \\ \hline
\end{tabular}%
}
\end{table}

As shown in Table \ref{tab:api}, similar awakening trends are observed on both DeepSeek-V3 and GPT4.1-mini. For example, \(B\)-\(R_x\)-\(A_x\) achieves a Hits@1 score of 16.1\% and a Hits@5 score (The API only provides 5 candidate answers) of 25.9\% on 2Wiki using DeepSeek, while \(A_q\)-instance-of and \(A_q\)-\(R_x\)-\(Z\)-\(R_x\)-\(B\) reach a Hits@1 score of 38.1\% and Hits@10 score of 45.0\% on CWQ with GPT4.1-mini. These results demonstrate that the awakening effect is not only effective on open-source LLMs, but also holds for API-based models.

\subsection{The Performance with Different Probing Method}
\label{probe}
As shown in Table \ref{tab:qa_nli}, we design a QA task by generating questions based on knowledge \(\mathcal{K}\); if the LLM can correctly answer the corresponding entity, it indicates that \(\mathcal{K}\) is embedded within the LLM. This setup evaluates the LLMs’ ability to perform complex utilization of knowledge, whereas the NLI task reflects a simpler form of knowledge usage. For example, the awakening knowledge \(A_q\)-\(R_q\)-\(B\), 110 samples were identified through QA probing, while 436 samples were identified through NLI probing, demonstrating that LLMs are more capable of handling simple knowledge usage. Under the same evaluation set (Intersection), the two probing methods detect comparable amounts of accessible knowledge.


\subsection{The Performance with Extra Awakening Knowledge}
\label{Extra_Awakening_Knowledge}
In the main text, we have explored the effects of six types of knowledge on awakening (T1, T2, T4, T5, T6, and T7). T3 is excluded from the main analysis due to its inconsistency with the reasoning direction, while T8 is omitted because the relations involved are weakly associated with the entities in the question. Therefore, we include T3 and T7 as additional experiments on CWQ dataset. As shown in Table \ref{tab:extra}, these results are consistent with our conclusion that even peripheral knowledge can successfully awaken the model to answer questions correctly.

\begin{table}[t]
\centering
\caption{The awakening performance of T3 and T7 across the six LLMs.}
\label{tab:extra}
\setlength{\tabcolsep}{0.07cm}{%
\begin{tabular}{ccccccc}
\toprule
\multirow{2}{*}{Knowledge Type} & \multicolumn{2}{c}{Qwen-7B} & \multicolumn{2}{c}{Llama-8B} & \multicolumn{2}{c}{Qwen-32B} \\ \cline{2-7} 
 & F & Hits@1 & F & Hits@1 & F & Hits@1 \\ \cline{2-7} 
T3 & 377 & 24.1 & 371 & 22.9 & 297 & 30.0 \\
T8 & 287 & 38.7 & 359 & 45.1 & 244 & 40.2 \\ \cline{2-7} 
\multicolumn{1}{l}{} & \multicolumn{2}{c}{Llama-70B} & \multicolumn{2}{c}{R1-70B} & \multicolumn{2}{c}{Qwen-72B} \\ \cline{2-7}
T3 & 380 & 31.6 & 201 & 31.8 & 307 & 25.4 \\
T8 & 287 & 38.0 & 152 & 48.0 & 246 & 35.8 \\ \bottomrule
\end{tabular}%
}
\end{table}

Furthermore, we conduct a comparative analysis of the impact of knowledge located at different hops on the model's awakening capability. The results show that knowledge closer to the correct answer tends to be more effective. For example, Qwen-7B-Instruct achieves a Hits@1 score of 38.7\% with second-hop knowledge \(A_x\)-\(R_q^2\)-\(B_x\), compared to only 8.4\% with first-hop knowledge \(A_x\)-\(R_q^1\)-\(B_x\). This suggests that semantically similar knowledge is often embedded in the same neurons.

\begin{table}[t]
\centering
\caption{Impact of the distance between awakening knowledge and answer.}
\label{tab:hops}
\setlength{\tabcolsep}{0.05cm}{%
\begin{tabular}{lcccccccccc}
\hline
\multicolumn{1}{c}{\multirow{2}{*}{Type}} & \multicolumn{2}{c}{Qwen-7B} & \multicolumn{2}{c}{Llama-8B} & \multicolumn{2}{c}{Qwen-32B} & \multicolumn{2}{c}{Llama-70B} & \multicolumn{2}{c}{Qwen-72B} \\ \cline{2-11} 
\multicolumn{1}{c}{} & \multicolumn{1}{l}{F} & \multicolumn{1}{l}{H@1} & \multicolumn{1}{l}{F} & \multicolumn{1}{l}{H@1} & \multicolumn{1}{l}{F} & \multicolumn{1}{l}{H@1} & \multicolumn{1}{l}{F} & \multicolumn{1}{l}{H@1} & \multicolumn{1}{l}{F} & \multicolumn{1}{l}{H@1} \\ \cline{2-11} 
\(A_x\)-\(R_q^1\)-\(B_x\) & 287 & 8.4 & 359 & 17.8 & 243 & 18.1 & 287 & 10.1 & 246 & 17.1 \\
\(A_x\)-\(R_q^2\)-\(B_x\) & 287 & 38.7 & 359 & 45.1 & 243 & 40.3 & 287 & 36.6 & 246 & 35.8 \\ \hline
\end{tabular}%
}
\end{table}

\subsection{The Complete Compare Results of Different Size LLMs.}
\label{size}
As shown in Table \ref{tab:size}, we compute the intersection of awakening results across different model sizes within the LLaMA and Qwen series for comparative analysis. The results indicate a general positive correlation between model size and awakening performance. However, for the Qwen series, the performance gain of the 72B model is limited, suggesting that awakening may also depend on architectural factors such as model depth. Figure \ref{tab:same_question} in the main text presents the average awakening scores of Qwen models across different knowledge types.

\begin{table*}[]
\centering
\caption{Impact of Llama and Qwen Model Sizes on Awakening Performance.}
\label{tab:size}
\setlength{\tabcolsep}{0.055cm}{%
\begin{tabular}{cccccccccccccccc}
\hline
\multirow{3}{*}{Knowledge Type} & \multicolumn{3}{c}{Llama-3.1-8B-Instruct} & \multicolumn{3}{c}{Llama-3.1-70B-Instruct} & \multicolumn{3}{c}{Qwen2.5-7B-Instruct} & \multicolumn{3}{c}{Qwen2.5-32B-Instruct} & \multicolumn{3}{c}{Qwen2.5-72B-Instruct} \\ \cline{2-16} 
 & F & H@1 & H@10 & F & H@1 & H@10 & F & H@1 & H@10 & F & H@1 & H@10 & F & H@1 & H@10 \\ \cline{2-16} 
 & \multicolumn{15}{c}{\textbf{2Wiki}} \\ \cline{2-16} 
\(A_q\)-\(R_q\)-\(B\) & 952 & 3.7 & 14.1 & 952 & 5.3 & 16.5 & 245 & 4.1 & 7.8 & 245 & 7.3 & 13.5 & 245 & 6.9 & 11.8 \\
\(A_q\)-\(R_x\)-\(B_x\) & 779 & 3.1 & 10.7 & 779 & 4.1 & 13.2 & 191 & 1.6 & 4.2 & 191 & 4.2 & 7.9 & 191 & 5.2 & 8.4 \\
\(B\)-\(R_x\)-\(A_x\) & 934 & 3.1 & 13.8 & 934 & 5.1 & 17.0 & 229 & 4.8 & 8.7 & 229 & 7.4 & 14.4 & 229 & 7.4 & 11.4 \\
\(A_x\)-\(R_q^1\)-\(B_x\) & 736 & 3.7 & 11.5 & 736 & 4.6 & 13.0 & 189 & 4.2 & 5.8 & 189 & 4.2 & 7.4 & 189 & 7.4 & 14.3 \\
\(A_q\)-\(R_x\)-\(Z\)-\(R_x\)-\(B\) & 478 & 4.0 & 11.7 & 478 & 5.0 & 16.1 & 55 & 0.0 & 5.5 & 55 & 0.0 & 7.3 & 55 & 9.1 & 12.7 \\
\(A_q\)-instance-of & 733 & 3.5 & 9.5 & 733 & 4.8 & 18.4 & 129 & 3.1 & 4.7 & 129 & 3.9 & 9.3 & 129 & 3.1 & 7.8 \\ \cline{2-16} 
\multicolumn{1}{l}{} & \multicolumn{15}{c}{\textbf{CWQ}} \\ \cline{2-16} 
\(A_q\)-\(R_q\)-\(B\) & 277 & 23.8 & 40.8 & 277 & 31.8 & 37.9 & 127 & 15.7 & 24.4 & 127 & 25.2 & 33.1 & 127 & 24.4 & 29.9 \\
\(A_q\)-\(R_x\)-\(B_x\) & 277 & 23.1 & 39.0 & 277 & 26.4 & 37.5 & 126 & 15.1 & 19.8 & 126 & 25.4 & 32.5 & 126 & 26.2 & 31 \\
\(B\)-\(R_x\)-\(A_x\) & 276 & 25.0 & 42.4 & 276 & 27.9 & 36.2 & 124 & 19.4 & 27.4 & 124 & 24.2 & 33.1 & 124 & 25 & 29.8 \\
\(A_x\)-\(R_q^1\)-\(B_x\) & 277 & 7.9 & 24.5 & 277 & 7.9 & 16.6 & 123 & 4.1 & 10.6 & 123 & 11.4 & 23.6 & 123 & 13.8 & 23.6 \\
\(A_q\)-\(R_x\)-\(Z\)-\(R_x\)-\(B\) & 270 & 18.9 & 39.6 & 270 & 28.1 & 37.0 & 101 & 10.9 & 16.8 & 101 & 19.8 & 29.7 & 101 & 25.7 & 27.7 \\
\(A_q\)-instance-of & 277 & 21.3 & 43.3 & 277 & 28.2 & 35.7 & 126 & 10.3 & 23.0 & 126 & 20.6 & 31.7 & 126 & 20.6 & 26.2 \\ \hline
\end{tabular}%
}
\end{table*}

\subsection{The Impact with Different Temperature in Awaken LLMs}
\label{temp}

\begin{figure}[t]
    \centering
    \includegraphics[scale=0.38]{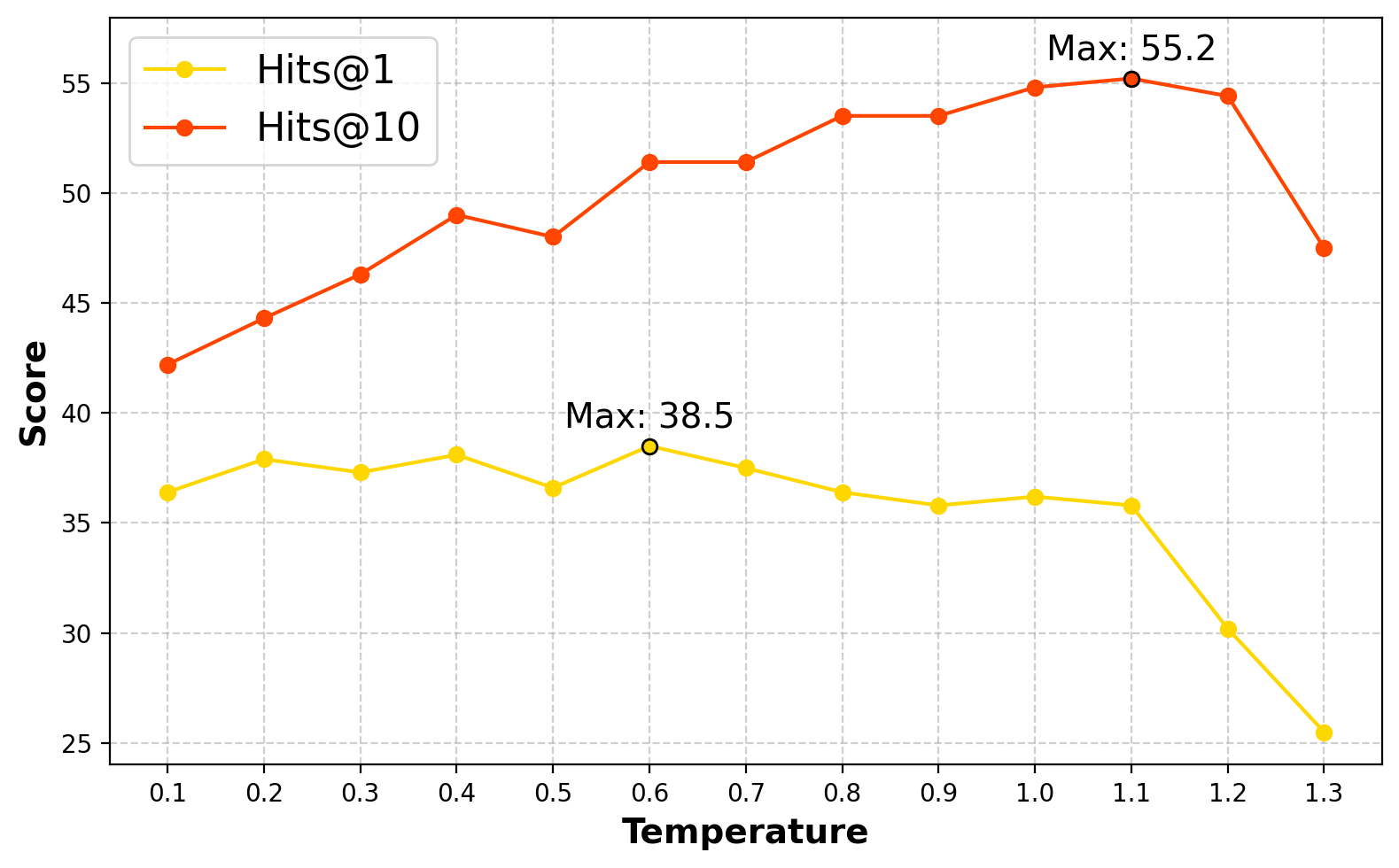}
    \caption{Impact of temperature on LLM Awakening with \(A_x\)-\(R_q\)-\(B_x\) on Llama3.1-8B-Instruct. Hits@1 and Hits@10 score get the best at 0.5 and 0.8, respectively.}
    \label{fig:temperature}
\end{figure}

Temperature also affects the ability of the model to awaken internal knowledge by controlling the randomness and determinism of the output of the LLM. As shown in Figure \ref{fig:temperature}, the performance first increases and achieving the best Hits@1 score of 18.8\% at a temperature of 0.5, and the best Hits@10 score at a temperature of 0.8. Additionally, a significant decline is observed once the temperature coefficient exceeds 1.0. This is because the distribution of the LLMs' output becomes smoother through the Softmax function, making low-probability tokens more likely to be selected.

\subsection{The Performance with \(A_q\)-\(R_q^1\) as Query}

In the main text, we use the full triplet as the query to select variant knowledge for evaluating awakening. However, in practice, entity \(B\)
is unobservable. Therefore, we also test using only the \(A_q\)-\(R_q^1\) as the query for knowledge selection. The results are shown in Table \ref{tab:a-r-similar}, the performance shows the similar trends with \(A_q\)-\(R_q^1\)-\(B\) as the query. For example, the awakening knowledge \(A_x\)-\(R_q^1\)-\(B_x\) achieves 2.8\% and 13.4\% on Hits@1 and Hits@10 scores with Llama3.1-8B-Instruct, respectively, which demonstrates the effectiveness of selecting knowledge based on entity \(A_q\) and relation \(R_q^1\) provides insightful conclusions for knowledge selection in the Unseen Entity QA task.

\begin{table*}[ht]
\centering
\caption{Performance of using \(A_q\)-\(R_q^1\) to select the top-1 most similar variant knowledge across five Instruction LLMs.}
\label{tab:a-r-similar}
\setlength{\tabcolsep}{0.078cm}{%
\begin{tabular}{cccccccccccccccc}
\toprule
\multicolumn{1}{c}{T} & \multicolumn{3}{c}{\textbf{Llama3.1-8B}} & \multicolumn{3}{c}{\textbf{Llama3.1-70B}} & \multicolumn{3}{c}{\textbf{Qwen2.5-7B}} & \multicolumn{3}{c}{\textbf{Qwen2.5-32B}} & \multicolumn{3}{c}{\textbf{Qwen2.5-72B}} \\ \cline{2-16} 
\multicolumn{1}{c}{} & F & Hits@1 & Hits@10 & F & Hits@1 & Hits@10 & F & Hits@1 & Hits@10 & F & Hits@1 & Hits@10 & F & Hits@1 & Hits@10 \rule{0pt}{2ex}\\ \cline{2-16} 
\multicolumn{1}{c}{} & \multicolumn{15}{c}{\textbf{2Wiki}} \rule{0pt}{2ex}\\ \cline{2-16} 
\rowcolor{red!10}
No RAG & 482 & 0.0 & 9.5 & 692 & 0.0 & 15.9 & 222 & 0.0 & 4.1 & 204 & 0.0 & 3.9 & 608 & 0.0 & 3.5 \rule{0pt}{2ex}\\
\rowcolor{green!10}
\textbf{\(A_q\)-\(R_q^1\)-\(B\)} & 482 & 4.6 & 16.4 & 692 & 5.6 & 17.3 & 222 & 3.6 & 6.3 & 204 & 6.4 & 14.2 & 608 & 7.1 & 11.2 \\
\rowcolor{green!10}
\textbf{\(A_q\)-\(R_x\)-\(B_x\)} & 482 & 3.7 & 12.2 & 692 & 4.9 & 15.5 & 222 & 3.2 & 6.3 & 204 & 3.9 & 13.2 & 608 & 5.9 & 10.2 \\
\rowcolor{green!10}
\textbf{\(B\)-\(R_x\)-\(A_x\)} & 482 & 3.3 & 14.1 & 692 & 5.3 & 19.4 & 222 & 3.6 & 6.3 & 204 & 4.4 & 15.2 & 608 & 7.7 & 13.5 \\
\rowcolor{green!10}
\textbf{\(A_x\)-\(R_q^1\)-\(B_x\)} & 482 & 2.1 & 14.5 & 692 & 4.6 & 14.7 & 222 & 4.5 & 5.9 & 204 & 6.4 & 12.3 & 608 & 6.9 & 12.2 \\
\rowcolor{green!10}
\textbf{\(A_q\)-\(R_x\)-\(Z\)-\(R_x\)-\(B\)} & 482 & 3.7 & 11.6 & 692 & 6.2 & 16.5 & 222 & 3.6 & 7.2 & 204 & 6.4 & 10.8 & 608 & 7.4 & 11.3 \\
\rowcolor{green!10}
\(A_q\)-\(\mathrm{instance}\)-\(\text{of}\) & 482 & 3.3 & 10.8 & 692 & 5.9 & 21.1 & 222 & 1.4 & 6.3 & 204 & 2.5 & 7.8 & 608 & 4.6 & 9.5 \\ \cline{2-16} 
 & \multicolumn{15}{c}{\textbf{CWQ}} \rule{0pt}{2ex}\\ \cline{2-16} 
 \rowcolor{red!10}
No RAG & 355 & 0.0 & 32.1 & 293 & 0.0 & 19.8 & 148 & 0.0 & 10.8 & 212 & 0.0 & 9.4 & 125 & 0.0 & 7.2 \rule{0pt}{2ex}\\
\rowcolor{green!10}
\textbf{\(A_q\)-\(R_q^1\)-\(B\)} & 355 & 39.7 & 55.8 & 293 & 33.8 & 42.3 & 148 & 17.6 & 28.4 & 212 & 37.7 & 46.2 & 125 & 20.8 & 28.0 \\
\rowcolor{green!10}
\textbf{\(A_q\)-\(R_x\)-\(B_x\)} & 355 & 31.5 & 56.1 & 293 & 27.6 & 44.4 & 148 & 18.9 & 30.4 & 212 & 27.4 & 46.2 & 125 & 23.2 & 28.8 \\
\rowcolor{green!10}
\textbf{\(B\)-\(R_x\)-\(A_x\)} & 355 & 37.2 & 56.9 & 293 & 27.6 & 43.0 & 148 & 18.9 & 30.4 & 212 & 33.5 & 45.3 & 125 & 24.0 & 26.4 \\
\rowcolor{green!10}
\textbf{\(A_x\)-\(R_q^1\)-\(B_x\)} & 355 & 18.0 & 39.2 & 293 & 8.9 & 23.9 & 148 & 5.4 & 11.5 & 212 & 17.0 & 29.2 & 125 & 17.6 & 26.4 \\
\rowcolor{green!10}
\textbf{\(A_q\)-\(R_x\)-\(Z\)-\(R_x\)-\(B\)} & 355 & 32.7 & 53.2 & 293 & 31.4 & 43.0 & 148 & 18.2 & 27.0 & 212 & 33.0 & 43.4 & 125 & 23.2 & 27.2 \\
\rowcolor{green!10}
\(A_q\)-\(\mathrm{instance}\)-\(\text{of}\) & 355 & 34.4 & 59.2 & 293 & 30.0 & 41.6 & 148 & 23.0 & 33.8 & 212 & 36.3 & 46.7 & 125 & 22.4 & 31.2 \\ \bottomrule
\end{tabular}%
}
\end{table*}

\section{Implementation Detail}
\label{app:exp}
For Awaken LLMs, we adopt VLLM to provide an OpenAI-compatible interface. All models are deployed using four NVIDIA A800 GPUs. For Unanswerable Question Selection and Knowledge Probing, the temperature parameter is randomly sampled from the range \([0,1]\). The temperature in Awaken Evaluation is set to 0.7. The values of Top-k and Top-p are kept in their default settings. In experiments involving Unseen Entity QA, for both entity-relation extraction and the awakening process, the temperature is fixed at 0.7, while Top-k and Top-p are also kept at their default settings.

The \(\mathrm{ACC_R}\) is defined as:
\begin{align}
    \operatorname{ACC_R}(\hat{y}, y) & = \left\{\begin{matrix}
  0, & \text {if} \forall g \in \text{ground\_list}, g \in \text{pre\_list} \\
  1, & \text { otherwise },
\end{matrix}\right.
\end{align}
where, \(\hat{y}\) is the predict result and \(y\) is gold label. \(\text{pre\_list}\) is the token list of the predicted answer, \(\text{ground\_list}\) is the list of gold label. Hits@1 indicates the accuracy of top-1 predictions, where correctness is determined using the \(\mathrm{ACC_R}\). Hits@10 follows the same evaluation protocol, measuring the proportion of correct answers within the top-10 predictions.

During the Incomplete KGs Construction, the limitation imposed by query time constraints necessitates capping the number of triples retrieved for constructing two-hop subgraphs to 40,000. This measure is implemented to maintain stability throughout the construction process.

During the Relation Retrieval phase, an Inverted File Index with Flat Quantization is employed to construct the index. The parameters used include an ``NLIST'' value of 4096, a ``TRAIN\_FACTO'' size of 40, and a ``SAMPLE\_PER\_BATCH'' quantity of 20,000.

\section{Datasets and Baselines}

\textbf{Datasets.} We evaluate our method on three widely used KGQA datasets: Mintaka, 2Wiki, and CWQ. Mintaka and 2Wiki are constructed based on Wikidata, a collaboratively edited knowledge graph that captures rich semantic relations between entities. Mintaka contains naturally elicited questions with diverse question types and high entity ambiguity, making it suitable for evaluating model generalization under realistic conditions. 2Wiki is designed to emphasize multi-hop reasoning and requires the model to traverse complex relational paths between entities. In contrast, CWQ is built on Freebase, a large-scale knowledge base with a different ontology. CWQ consists of complex questions automatically generated from web search queries and rewritten by crowdworkers to improve naturalness. It serves as a benchmark for evaluating a model's ability to handle compositionality and implicit constraints.

As shown in Table \ref{tab:dataset}, The experiments in this paper are conducted on three KGQA datasets: Mintaka (4000 samples), 2Wiki (3000 samples), and CWQ (1000 samples). Among them, 2Wiki and CWQ are further filtered to ensure the presence of gold reasoning paths required for knowledge awakening.

\begin{table}[t]
\centering
\caption{Statistics of Evaluation Datasets. Since Freebase may lack certain entity names involved in the reasoning process, we select questions with complete reasoning path information for evaluation in CWQ.}
\label{tab:dataset}
\setlength{\tabcolsep}{1.4cm}{
\begin{tabular}{cc}
\toprule
Datasets & Number \\ \midrule
Mintaka & 4000 \\ 
2Wiki & 3000 \\
CWQ & 1000 \\ \bottomrule
\end{tabular}%
}
\end{table}

\textbf{Baselines.} \textbf{DiFaR.} In this paper, we build several variants based on the core DiFaR method; therefore, we primarily describe the technical pipeline of DiFaR. DiFaR was proposed to alleviate the answer errors in existing KGQA methods caused by entity linking failures. It transforms the entire KG triples into sequences, which are then encoded using an embedding model and retrieved following the traditional RAG framework. DiFaR provides both unsupervised and supervised settings; however, due to the large size of our models (over 32B), the supervised setup is impractical. Therefore, we focus on the unsupervised setting in this paper.

\textbf{QD-DiFaR.} Question decomposition has been proven effective in addressing complex multi-hop questions. By breaking down a complex query into a sequence of simpler single-hop sub-questions for answering or evidence retrieval, it allows for better utilization of both the question context and the internal knowledge of language models. Therefore, we combine question decomposition with the DiFaR approach as a baseline \textbf{QD-DiFaR}.

\section{The Prompt Template Used in Awakening and Unseen Entity QA}
\label{app:prompt}

\begin{tcolorbox}[colback=white!5!white, colframe=black!75!black, title=The prompt template used in Unanswerable Question select]

Please answer the question, and put the short answer after the '- Short Answer:' \\
Here are some examples: \\
Example1: \\
- Question: What is the capital of France? \\
- Short Answer: Paris. \\
Please follow the above examples format to inference a new one: \\
- Question: {} \\
- Short Answer:
\end{tcolorbox}

\begin{tcolorbox}[colback=white!5!white, colframe=black!75!black, title=The prompt template used in Knowledge Probing]

Determine the truthfulness of a given claim using logical reasoning and evidence-based evaluation. \\
- Analyze the claim and break it down into its components.\\
- Identify any keywords or concepts that are crucial to understanding the claim.\\
**Example 1:**\\
- **Claim**: "The Great Wall of China is visible from space."\\
- **Conclusion**: False. While large, the Great Wall of China is not visible from space without aid. \\
- Consider any contextual factors that may influence the claim's truthfulness. Do not generate the examples, only inference the new one.\\
**Please inference new one:**\\
- **Claim**: {}\\
- **Conclusion**:
\end{tcolorbox}

\begin{tcolorbox}[colback=white!5!white, colframe=black!75!black, title=The prompt template used in Awakening LLMs]

I will give you some knowledge about the question, the knowledge may be not help you directlt answer the question, but it may be can help you to awake the knowledge in your parameters, you need to combine the knowledge which pre-trained to answer the question. \\
Here are some examples: \\
Example:\\
- Knowledge: Paris is the capital of France. \\
- Question: What is the capital of France?\\
- Short Answer: Paris.\\
Please follow the above examples format to inference a new one:\\
- Knowledge: {}\\
- Question: {}
\end{tcolorbox}

\begin{tcolorbox}[colback=white!5!white, colframe=black!75!black, title=The prompt template used in Entity-Relation Extraction]
Task: Extract the relation, in Wikidata style, present in a multi-hop query in relation to a given entity. \\
\# Task Output Format\\
- Provide the output in the format: **Output:** entity, relation.\\
\# Task Examples\\
- **Input:** Query: "What is the profession of the sibling of Albert Einstein?" Entity: "Albert Einstein"\\
  **Output:** "Albert Einstein, sibling"\\
 Now, you need extract the relation from the query according to the entity.\\
 - **Input:** Query:"{}" Entity:"{}"
\end{tcolorbox}

\end{document}